\documentclass[11pt]{article}

\usepackage[margin=1in]{geometry}
\usepackage{microtype}
\usepackage{setspace}
\usepackage{amsmath,amssymb,amsfonts,amsthm}
\usepackage{mathtools}
\usepackage{mathrsfs}
\usepackage{physics}

\usepackage{graphicx}
\usepackage{booktabs}
\usepackage{multirow}
\usepackage{makecell}
\usepackage{caption}
\usepackage{subcaption}

\usepackage{algorithm}
\usepackage{algpseudocode}

\usepackage{enumitem}
\usepackage[usenames,dvipsnames]{xcolor}

\usepackage{natbib}
\usepackage[
    colorlinks=true,
    linkcolor=MidnightBlue,
    citecolor=MidnightBlue,
    urlcolor=MidnightBlue,
    pdftitle={Representative Learning for Distributed Support Vector Machines},
    pdfauthor={Joseph A. Casey, Boxiang Wang, Keren Li}
]{hyperref}

\usepackage{authblk}

\title{\textbf{Representative Learning for Distributed Support Vector Machines}}

\author[1]{Joseph A. Casey}
\author[2]{Boxiang Wang}
\author[1]{Keren Li\thanks{Corresponding author: \href{mailto:kli@uab.edu}{kli@uab.edu}}}

\affil[1]{Department of Mathematics, University of Alabama at Birmingham}
\affil[2]{Department of Statistics and Actuarial Science, University of Iowa}

\date{}

\newtheorem{proposition}{Proposition}
\newtheorem{lemma}{Lemma}
\newtheorem{theorem}{Theorem}
\newtheorem{corollary}{Corollary}
\theoremstyle{remark}
\newtheorem{remark}{Remark}

\begin{document}

\maketitle

\begin{abstract}{}
Distributed support vector machine (SVM) learning is challenging under heterogeneous local data and limited communication, particularly because the nonsmooth hinge loss complicates gradient-based optimization and the construction of compact summaries of local objectives. We develop a Representative Learning framework based on Lifted Empirical Risk Minimization, in which each covariate is replaced by a local Gaussian distribution. For linear SVMs, the resulting expected hinge loss has an exact closed form that is smooth for nondegenerate lifting covariances and converges to the ordinary hinge loss as the covariance vanishes. 

The lifted formulation leads naturally to clouded representatives that encode both location and covariance information. We study a clouded Mean Representative (MR), which summarizes class-wise local data, and a clouded Score-Matching Representative (SMR), which additionally matches the class-wise gradient of the lifted objective at its construction parameter. For SMR, we show that the representative covariance scale remains close to the original scale as within-group covariate heterogeneity decreases, and that the discrepancy between the full-data gradient and the gradient evaluated on frozen representatives is controlled by the largest within-group diameter. Under strong convexity and smoothness, this yields an inexact-gradient convergence guarantee whose error floor is proportional to the representative approximation error.

Synthetic experiments under heterogeneous class allocation show that the lifted representative methods remain competitive with centralized solutions when parameter-aggregation methods are affected by repeated local optimization. Although MR can achieve similar predictive accuracy, SMR more closely recovers the centralized lifted-SVM optimizer. We further extend the framework to nonlinear SVMs through a shared kernel dictionary that provides a common finite-dimensional coordinate system for constructing and communicating representatives. Experiments on the Ionosphere and Australian Credit Approval datasets show analogous behavior in the kernel setting. Together, these results provide a distributional representative framework for communication-efficient distributed large-margin learning under heterogeneity.
\end{abstract}


\newpage

\section{Introduction}\label{sec:intro}

Distributed Machine Learning (DML) enables learning from data distributed across multiple nodes without requiring the raw data to be centralized. Federated Learning (FL) provides a prominent framework for this setting, in which models are trained locally and information is communicated across nodes or through a coordinating server \citep{mcmahan2017communication,kairouz2021advances}. Despite substantial progress, distributed learning remains challenging when data are heterogeneous and communication is constrained. Local datasets are often non-IID, with potentially substantial differences in feature and response distributions, so locally trained models or updates may provide conflicting information about the global learning objective. At the same time, privacy and data-localization requirements can prohibit direct exchange of raw observations, while limited bandwidth, intermittent connectivity, and asynchronous participation restrict the frequency and volume of communication. Consequently, a fundamental question in DML is not only how to optimize a learning objective across distributed data, but also what information should be communicated between nodes to support accurate and robust learning under these constraints.

Support Vector Machines (SVMs) provide an important setting for studying distributed classification. As classical margin-based classifiers, SVMs offer strong predictive performance, a convex training objective, and flexible nonlinear decision boundaries through kernel methods \citep{cortes1995support,vapnik1999overview}. Their relevance also extends beyond classical kernel learning. Modern neural networks with rectified linear unit (ReLU) activations construct complex piecewise-linear prediction functions, creating useful conceptual connections between margin-based learning and modern deep architectures. Thus, SVMs provide a mathematically tractable setting in which the effects of data geometry, margins, and distributed information sharing can be studied explicitly, while retaining connections to broader classes of modern learning models.

At the same time, SVMs pose distinctive challenges in distributed environments. Their empirical objective is based on the nonsmooth hinge loss, and the learned decision boundary depends critically on observations near or within the margin. When data are distributed across heterogeneous nodes, the observations that are locally important for defining a decision boundary may differ substantially across nodes. One natural strategy is therefore to communicate support vectors, which characterize the SVM solution, but support vectors are themselves observations from the local datasets and may constitute a substantial portion of the data \cite{graf2004parallel}. Other distributed SVM approaches coordinate local classifiers through repeated exchange of model parameters or consensus updates, for example through ADMM-based formulations \citep{forero2010consensus,zhang2018consensus}. Kernel SVMs introduce an additional difficulty because nonlinear decision functions may be represented through data-dependent kernel expansions, motivating approximations such as Nystr\"om methods or Random Fourier Features when explicit finite-dimensional representations are needed \cite{williams2000using,rahimi2007random}. These characteristics make distributed SVM learning a useful testbed for asking how the information needed to determine a global decision boundary can be communicated compactly without directly transmitting the underlying observations or relying on frequent model exchange.

Representative Learning (RepL) provides an alternative approach to distributed communication by replacing raw observations or repeated model updates with compact pseudo-data representatives constructed from local datasets \citep{li2022score,zheng2024rasmr,li2026asmr}. Existing RepL methods have primarily focused on point-valued representatives, including mean- and score-based constructions that preserve selected statistical or optimization information from local data. While effective in many settings, point representatives may not adequately capture local variability and geometric information that can be particularly important for nonsmooth learning problems.

To address this limitation, \cite{wang2026quantile} introduced \emph{clouded representatives}, extending RepL from point-valued to distribution-valued representatives, together with a Lifted Empirical Risk Minimization (Lifted ERM) framework in which individual observations are replaced by local probability measures and learning proceeds through expected loss over these distributions. Classical ERM is recovered as the special case in which the probability measures degenerate to point masses. Originally developed for distributed quantile regression, this framework provides a natural foundation for extending RepL to other nonsmooth learning problems. In this work, we build on the Lifted ERM and clouded-representative framework of \cite{wang2026quantile} to develop representative learning for distributed SVMs with hinge loss.

Our contributions are fourfold.
\begin{enumerate}[label=(\roman*)]
    \item We formulate a lifted empirical risk for SVMs in which observations are represented by local probability distributions and derive the resulting lifted hinge loss for Gaussian clouds.
    \item We construct clouded representatives for distributed linear SVMs and characterize their approximation to the corresponding local empirical risks/gradients.
    \item We extend the framework to nonlinear classification through RKHS representations and kernel approximations.
    \item We evaluate the resulting distributed algorithms under heterogeneous node distributions using synthetic and benchmark datasets.
\end{enumerate}

\section{Background and Literature Review}\label{sec:bg} 

DML considers learning from data partitioned across multiple computational nodes, with distributed optimization methods ranging from parallel stochastic gradient descent and parameter-server architectures to modern FL \citep{zinkevich2010parallelized, li2014scaling, konecny2016federated}. FL emphasizes settings in which data are naturally distributed across devices or institutions and raw observations remain local, with Federated Averaging (FedAvg) providing a canonical approach based on alternating local optimization and global parameter aggregation \citep{mcmahan2017communication,kairouz2021advances}. A central challenge is that local data are generally non-IID, so repeated local optimization can introduce client drift and degrade convergence or statistical accuracy \citep{zhou2018convergence,haddadpour2019convergence,khaled2020tighter,koloskova2020unified,li2020federated}. A substantial literature consequently modifies local or global optimization to improve robustness to heterogeneity, including proximal and variance-corrected methods, local-update and communication-efficient SGD, and robust aggregation or sampling strategies \citep{li2020federated,karimireddy2020scaffold,stich2018local,stich2020error,spiridonoff2021communication,wang2021cooperative,wang2021device}. Related approaches relax the requirement of a single global model through personalized and multi-task federated learning \citep{fallah2020personalized,li2021ditto,tan2022towards,smith2017federated,argyriou2006multi,zhang2012convex}, while decentralized methods replace centralized aggregation by peer-to-peer communication \citep{lian2017decentralized,assran2019stochastic}. Another line reduces dependence on parameter exchange through knowledge distillation, communicating model predictions rather than complete model states and thereby accommodating, in some settings, heterogeneous local architectures \citep{hinton2015distilling,chang2019cronus,lin2020ensemble,wu2022communication,zhu2021datafree}. Distributionally robust optimization provides a complementary approach by optimizing against distributional perturbations to improve robustness under distribution shift \citep{kuhn2019wasserstein}. Despite these developments, most distributed learning methods communicate quantities derived from the current model, such as parameters, gradients, or predictions, and their effectiveness can remain sensitive to data heterogeneity, communication frequency, model compatibility, and synchronization. These limitations motivate communication mechanisms that summarize the local data itself rather than only the state or output of a local model.

Data heterogeneity is one of the central challenges in distributed learning because local datasets are rarely independent and identically distributed across nodes. Heterogeneity can arise through differences in feature distributions $P_k(X)$, label distributions $P_k(Y)$, conditional relationships $P_k(Y|X)$, or class-conditional feature distributions $P_k(X|Y)$, corresponding broadly to feature, label, model, and representation heterogeneity. These differences cause local objectives and local update directions to deviate from the global learning objective, so algorithms based on repeated local optimization and parameter aggregation can suffer from client drift, slower convergence, instability, and degraded global performance \cite{zhou2018convergence,haddadpour2019convergence,khaled2020tighter,koloskova2020unified,li2020federated,karimireddy2020scaffold}. The problem becomes particularly pronounced when nodes have severe class imbalance, distribution shift, or substantially different local prediction relationships. A broad literature has therefore sought to mitigate heterogeneity through proximal or control-variate corrections, robust aggregation, personalized learning, multi-task learning, and distributionally robust optimization. More generally, heterogeneity changes the goal of distributed learning from simply reproducing centralized optimization to determining how information from statistically different nodes should be combined so that collaboration remains beneficial.

\subsection{Representative Learning: From Point to Cloud}\label{sec:bg_repl}

RepL provides a data-centric framework for distributed learning in which nodes communicate compact pseudo-data, called \emph{representatives}, rather than repeatedly exchanging model parameters or gradients \cite{li2022score,li2026asmr}. 

Suppose the full dataset is distributed across $K$ nodes, 
\begin{align*}
    \mathcal{D} = \cup_{k=1}^{K}\mathcal{D}_k, \quad \mathcal{D}_k = \{(\mathbf{x}_i, y_i)_{i \in I_k}\},
\end{align*}
where $I_k$ is the index set for node $k$ and $n_k=|I_k|$ is its sample size. A weighted point representative of $\mathcal D_k$ takes the form $\tilde{\mathcal{D}}_k = \left(n_k,\tilde{\mathbf{x}}_k, \tilde{y}_k \right)$, constructed from the local data.

More generally, a node may be divided into $r_k$ subpartitions, giving the representative dataset
\begin{align*}
    \tilde{\mathcal{D}}_k = \left\{(n_{kh},\tilde{\mathbf{x}}_{kh}, \tilde{y}_{kh})_{h=1}^{r_k} \right\}
\end{align*}

The collection $\tilde{\mathcal D} =\bigcup_{k=1}^K \tilde{\mathcal D}_k$ then provides a compact pseudo-dataset on which downstream learning can be performed. Different representative constructions differ primarily in which statistical or optimization information of the original data they seek to preserve.

The earliest RepL constructions are point-valued representatives. The \emph{Mean Representative} (MR) provides a model-free summary by taking the coordinatewise means of the observations within a local group \cite{li2022score}. MR is particularly natural for linear models, but direct averaging does not generally preserve the nonlinear relationship between the response and the linear predictor in generalized linear models (GLMs). To better account for this nonlinear structure, the \emph{Transformed Mean Representative} (TMR) constructs the response representative in the transformed space \cite{li2026asmr}. TMR therefore retains the simplicity of a mean-based representative while incorporating model-specific transformation information.

A different construction principle is provided by the \emph{Score-Matching Representative} (SMR), which constructs representatives to match local score function values  in a GLM \cite{li2022score}. Thus, unlike MR and TMR, SMR is explicitly model-oriented and encodes information from the local score function into the representative. The \emph{Response-Aided Score-Matching Representative} (RASMR) \cite{zheng2024rasmr} refines heterogeneous local blocks using response- and model-related information, to improve the construction of representative points. The \emph{Anchored Score-Matching Representative} (Anchored-SMR) addresses the original underdetermined score-matching system by matching the intercept component of the score. Anchored-SMR provides an identifiable score-matching construction with a fixed intercept representation. 

Despite these developments, the representative constructions above remain point-valued: a collection of observations is ultimately summarized by one or more weighted pseudo-observations. Compression to isolated points can discard information about within-group variability and local geometry, which may be particularly important for nonsmooth learning objectives.

To incorporate such information, \cite{wang2026quantile} introduced \emph{Lifted Empirical Risk Minimization} (Lifted ERM) and \emph{clouded representatives}, extending point-based RepL to distribution-valued observations. Let $\ell(f_\theta(\mathbf{x}),y)$ denote a loss function evaluated under a predictive model $f_\theta$. Each observed covariate $\mathbf{x}_i$ is associated with a local probability measure $Q_i$, and the lifted empirical risk is defined as
\begin{align} 
    R_n^Q(\theta) = \frac{1}{n} \sum_{i=1}^n \mathbb E_{\mathbf{X}_i\sim Q_i} \left[ \ell\left(f_\theta(\mathbf{X}_i),y_i\right) \right]. \label{eq:lifted_erm} 
\end{align}
When $Q_i=\delta_{\mathbf{x}_i}$, with $\delta_{\mathbf{x}_i}$ denoting the Dirac measure at $\mathbf{x}_i$, equation \eqref{eq:lifted_erm} reduces to the classical empirical risk. Lifted ERM therefore embeds conventional point-based ERM as a special case while permitting learning over local probability distributions rather than isolated observations \cite{wang2026quantile}.

Within this framework, \cite{wang2026quantile} introduced a \emph{clouded representative} as a distributional extension of a point representative. Rather than representing a local group only through a response and covariate location, a clouded pseudo-sample additionally carries distributional information describing local variability. For distributed quantile regression, \cite{wang2026quantile} develops clouded representatives containing representative response, location, and covariance information, and constructs Clouded SMR by matching score equations under the lifted risk. This distribution-valued representation preserves local geometric and variability information that is absent from a point summary and is especially relevant for nonsmooth objectives, where expected loss can depend on how probability mass is distributed around the nondifferentiable region of the loss. 

\subsection{Distributed Support Vector Machines}\label{sec:bg_dsvm}

For observations $(\mathbf{x}_i,y_i)$, with $\mathbf{x}_i\in\mathbb{R}^p$ and $y_i\in\{-1,1\}$, a linear classifier takes the form $f(\mathbf{x})=\mathbf{w}^\top\mathbf{x}+b$. The soft-margin SVM can be expressed through the regularized hinge-loss problem 
\begin{align} 
    \min_{\mathbf{w},b}\; \frac{\lambda}{2}\|\mathbf{w}\|_2^2 + \frac{1}{n}\sum_{i=1}^n \max\left\{ 0, 1-y_i(\mathbf{w}^\top\mathbf{x}_i+b) \right\}. 
    \label{eq:svm_erm} 
\end{align} 
The hinge loss is convex but nonsmooth at observations satisfying $y_i(\mathbf{w}^\top\mathbf{x}_i+b)=1$. This nonsmooth structure distinguishes SVM optimization from smooth likelihood-based problems and commonly motivates subgradient, coordinate-descent, or quadratic-programming methods.

In distributed settings, if the objective is to recover a single global classifier, the centralized SVM objective can be decomposed into node-level hinge-loss contributions,
\begin{align} 
    \min_{\mathbf{w},b}\; \frac{\lambda}{2}\|\mathbf{w}\|_2^2 + \frac{1}{n} \sum_{k=1}^K \sum_{i\in I_k} \max\left\{ 0, 1-y_i(\mathbf{w}^\top\mathbf{x}_i+b) \right\}. 
    \label{eq:dsvm_global} 
\end{align}
The statistical target is therefore well defined even when the observations cannot be pooled. The central difficulty is how to solve \eqref{eq:dsvm_global} when each node has access only to its local data and communication between nodes is restricted.

Existing distributed SVM methods approach this problem through different choices of communicated information. One natural strategy is to exchange support vectors. From the dual representation, $\mathbf{w} = \sum_{i=1}^n \alpha_i y_i\mathbf{x}_i$,  only observations with $\alpha_i>0$ contribute directly to the classifier. These support vectors therefore provide a natural data-based summary of local SVM solutions and have been used in hierarchical distributed training schemes. However, support vectors are observations from the original local datasets, and their number need not be small. Consequently, communicating support vectors may conflict with privacy or communication restrictions in distributed learning. Related approaches use coresets or other reduced subsets to decrease the amount of data needed for SVM training.

A second class of methods communicates model parameters and enforces agreement between local SVMs. For example, consensus-based distributed SVM can assign a local classifier $(\mathbf{w}_k,b_k)$ to each node and solve 
\begin{align} 
    \min_{\{\mathbf{w}_k,b_k\}_{k=1}^K}\; &\sum_{k=1}^K \left[ \frac{1}{2}\|\mathbf{w}_k\|_2^2 + C\sum_{i\in I_k} \max\left\{ 0, 1-y_i(\mathbf{w}_k^\top\mathbf{x}_i+b_k) \right\} \right] \label{eq:dsvm-consensus}\\ 
    \text{subject to}\; &\mathbf{w}_1=\cdots=\mathbf{w}_K, \notag\\
    &b_1=\cdots=b_K. \notag 
\end{align} 
Such formulations can be solved by alternating local SVM optimization with consensus updates, for example through ADMM. Rather than exchanging local observations, communication is shifted to model parameters and consensus variables.

SVMs can also be incorporated into more general distributed or federated optimization frameworks. Local nodes may optimize the hinge loss using subgradient methods and periodically aggregate model parameters, while proximal penalties can be introduced to restrict local models from drifting too far from a shared model. These approaches avoid direct exchange of training observations, but their communication remains model-based. When local datasets are heterogeneous, multiple local optimization steps can move node-specific models in different directions before communication, creating the same local-drift issue that arises more generally in distributed optimization.

These approaches illustrate a broader design question in distributed SVMs: what information should be communicated so that decentralized optimization remains close to the centralized SVM objective? Support vector methods communicate selected observations, consensus and federated methods communicate model information, and coreset methods communicate reduced subsets of the data. Each choice determines which aspects of the local learning problem are preserved during communication. For SVMs, this question is particularly important because the hinge loss is nonsmooth and depends critically on the distribution of observations near the decision margin. A compact distributed representation should therefore capture not only the location of local observations but also the information relevant to their contribution to the SVM objective. This motivates the development of representative learning methods specifically adapted to distributed large-margin classification.

\newpage

\section{Representative Learning for Distributed SVM}\label{sec:dvm}

We develop RepL methods for training SVMs from data distributed across heterogeneous nodes. Building on the Lifted ERM and clouded-representative framework introduced in \cite{wang2026quantile}, we replace each point-valued covariate by a local probability distribution and evaluate the hinge loss through its expectation over that distribution. For Gaussian clouds, the resulting lifted hinge loss has a closed-form, differentiable expression that incorporates both the location and covariance of the cloud. This formulation enables gradient-based optimization and provides the foundation for constructing class-wise clouded representatives carrying local geometric and optimization information.

We first develop the method for linear SVMs. We derive the Gaussian lifted hinge loss, establish its principal analytical properties, and construct clouded MR and SMR. These representatives are communicated between nodes and incorporated into the local learning objectives to reduce the effects of data heterogeneity and local model drift without exchanging raw observations, gradients, or model parameters. We then extend the framework to kernel SVMs, where representatives lie in a reproducing kernel Hilbert space and require a common representation that can be communicated across nodes.

\subsection{Linear SVM}

\subsubsection{Lifted SVM} 

To obtain a smooth objective while retaining information about local variability, we formulate the SVM through Lifted ERM, in which each point-valued covariate is replaced by a probability distribution. Specifically, we lift each observation $\mathbf{x}_i$ to a Gaussian cloud
\begin{align}\label{eq:svm_gaussian_lift} 
    \tilde{\mathbf{X}}_i \sim N(\mathbf{x}_i,\boldsymbol{\Sigma}), 
\end{align}
where $\boldsymbol{\Sigma} = \operatorname{diag} \left( \tilde{\sigma}_1^2,\ldots, \tilde{\sigma}_p^2 \right)$. A convenient default choice is the coordinate-wise rule-of-thumb bandwidth $\tilde{\sigma}_j = \widehat{\sigma}_j \left(\frac{3n}{4}\right)^{-1/5}$, $j=1,\ldots,p$, with $\widehat{\sigma}_j$ the sample standard deviation of the $j$th covariate. This choice follows the classical rule-of-thumb bandwidth used in Gaussian density estimation. More generally, $\boldsymbol{\Sigma}$ controls the scale and geometry of the lifting and need not be restricted to a diagonal form.

Under the lifting in \eqref{eq:svm_gaussian_lift}, the hinge-loss contribution of observation $i$ becomes 
\begin{align}\label{eq:lifted_hinge_definition} 
    \tilde{\ell}_i(\mathbf{w},b) = \mathbb{E}_{\tilde{\mathbf{X}}_i} \left[ \max\left\{ 0, 1-y_i (\mathbf{w}^{\top}\tilde{\mathbf{X}}_i+b) \right\} \right]. 
\end{align} 
Consequently, the lifted SVM solves 
\begin{align}\label{eq:lifted_svm} 
    \min_{\mathbf{w},b} \frac{1}{2}\|\mathbf{w}\|_2^2 + C\sum_{i=1}^n \tilde{\ell}_i(\mathbf{w},b). 
\end{align}

The expectation in \eqref{eq:lifted_hinge_definition} has an exact closed form. Define the random margin 
\begin{align*}  
    M_i = y_i (\mathbf{w}^{\top}\tilde{\mathbf{X}}_i+b). 
\end{align*}
Since $\tilde{\mathbf{X}}_i$ is Gaussian, 
\begin{align*} 
    M_i \sim N(\nu_i,s^2), 
\end{align*}
where 
\begin{align}\label{eq:margin_mean_var} 
    \nu_i = y_i(\mathbf{w}^{\top}\mathbf{x}_i+b), \quad s^2 = \mathbf{w}^{\top} \boldsymbol{\Sigma} \mathbf{w}. 
\end{align} 
Let \[ a_i=\frac{1-\nu_i}{s}, \] and denote the standard normal distribution and density functions by $\Phi$ and $\phi$, respectively. For $s>0$, 
\begin{align}\label{eq:closed_form_lifted_hinge} 
    \tilde{\ell}_i(\mathbf{w},b) = (1-\nu_i)\Phi(a_i) + s\phi(a_i). 
\end{align} 
Thus, rather than replacing the hinge loss by an unrelated smooth surrogate, the lifted formulation evaluates the \emph{exact expected hinge loss} under the Gaussian cloud.

\begin{figure}[t] 
\centering 
\includegraphics[width=0.60\textwidth] {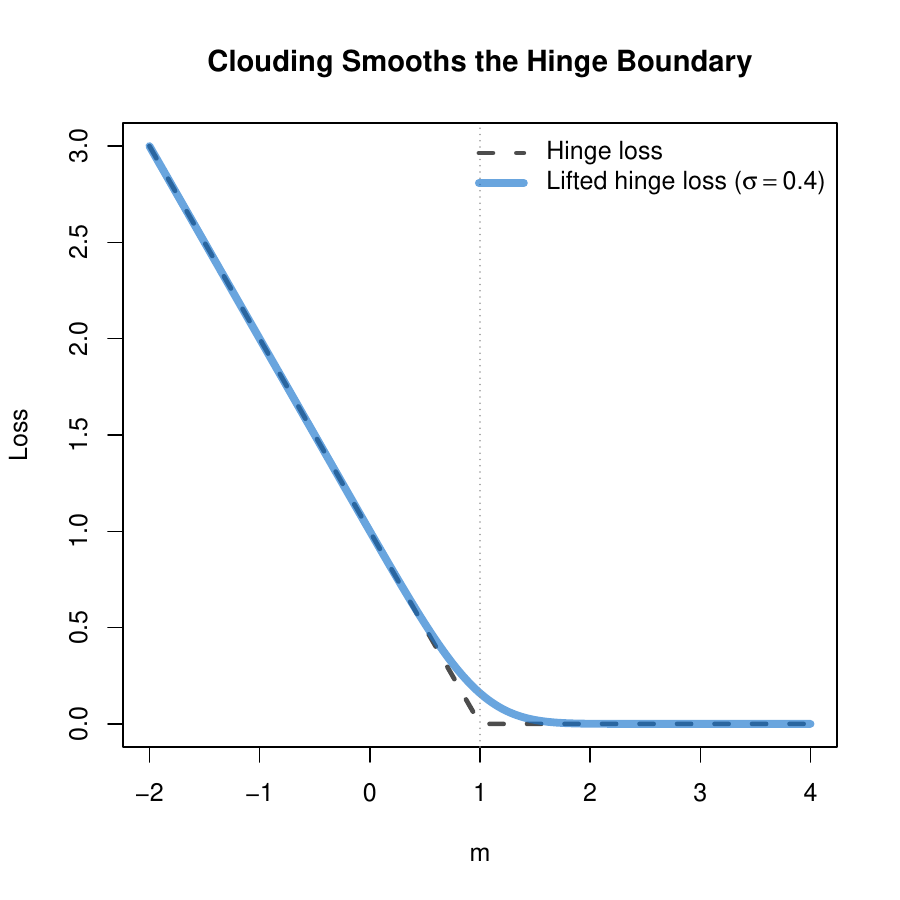} 
\caption{Hinge loss vs lifted hinge loss.} 
\label{fig:hinge_lifted} 
\end{figure}

Figure~\ref{fig:hinge_lifted} illustrates the smoothing induced by Gaussian lifting around the nondifferentiable margin boundary.

For a single observation, write
\begin{align}\label{eq:lifted_hinge_generic}
    \ell_{\Sigma}(\mathbf{w},b) = (1-\nu)\Phi(a)+s\phi(a).
\end{align}
Unless otherwise stated, assume $\boldsymbol{\Sigma}\succeq0$. For results involving derivatives, we further restrict attention to $s\geq\epsilon>0$.

\begin{proposition}[Smoothness of the lifted hinge loss] \label{prop:lifted_smoothness} 
Suppose that $s=\sqrt{\mathbf{w}^\top\boldsymbol{\Sigma} \mathbf{w}}\geq\epsilon>0$ on the parameter set under consideration and that $\mathbf{x}$ and $\boldsymbol{\Sigma}$ are bounded. Then $\ell_{\Sigma}(\mathbf{w},b)$ has a Lipschitz-continuous gradient on this set. Consequently, $\ell_{\Sigma}$ is $L$-smooth for some finite $L>0$. 
\end{proposition}

\begin{proof} 
Differentiating \eqref{eq:lifted_hinge_generic} gives 
\begin{align} 
    \nabla_w \ell_{\Sigma}(\mathbf{w},b) &= -y\Phi(a)x + \phi(a)\frac{\boldsymbol{\Sigma} \mathbf{w}}{s}, \label{eq:lifted_gradient_w}\\ 
    \frac{\partial\ell_{\Sigma}(\mathbf{w},b)}{\partial b} &= -y\Phi(a). \label{eq:lifted_gradient_b} 
\end{align} 
The corresponding second derivatives are 
\begin{align} 
    \nabla_w^2\ell_{\Sigma}(\mathbf{w},b) &= \frac{\phi(a)}{s} \left( y\mathbf{x}+a\frac{\boldsymbol{\Sigma} \mathbf{w}}{s} \right) \left( y\mathbf{x}+a\frac{\boldsymbol{\Sigma} \mathbf{w}}{s} \right)^\top \nonumber\\ &\quad+ \phi(a) \left\{ \frac{\Sigma}{s} - \frac{(\boldsymbol{\Sigma} \mathbf{w})(\boldsymbol{\Sigma} \mathbf{w})^\top}{s^3} \right\}, \label{eq:lifted_hessian_w}\\ 
    \nabla_w\frac{\partial\ell_{\Sigma}}{\partial b} &= \frac{\phi(a)}{s} \left( y\mathbf{x}+a\frac{\boldsymbol{\Sigma} \mathbf{w}}{s} \right), \label{eq:lifted_hessian_wb}\\ \frac{\partial^2\ell_{\Sigma}}{\partial b^2} &= \frac{\phi(a)}{s}. \label{eq:lifted_hessian_b} 
\end{align} 
Since $\phi(a)\leq(2\pi)^{-1/2}$, $s\geq\epsilon>0$, and the remaining quantities are bounded on the parameter set, the Hessian has bounded operator norm. Hence the gradient is Lipschitz continuous. 
\end{proof} 

\begin{proposition}[Convexity and regularized strong convexity] \label{prop:lifted_convexity} 
The lifted hinge loss $\ell_{\Sigma}(\mathbf{w},b)$ is convex in $(\mathbf{w},b)$. Moreover, for $\lambda>0$, adding $\lambda\|\mathbf{w}\|^2/2$ makes the resulting objective $\lambda$-strongly convex in $\mathbf{w}$ for each fixed $b$. 
\end{proposition} 

\begin{proof}
Using the expectation representation, 
\[ 
\ell_{\Sigma}(\mathbf{w},b) = \mathbb{E}_{\xi\sim N(0,\boldsymbol{\Sigma})} \left[ \max\left\{ 0, 1-y\bigl(\mathbf{w}^\top(\mathbf{x}+\xi)+b\bigr) \right\} \right]. 
\] 
For each fixed $\xi$, the expression inside the expectation is the maximum of two convex functions of $(\mathbf{w},b)$ and is therefore convex. Expectation preserves convexity, proving the first claim. Since $\lambda\|\mathbf{w}\|^2/2$ is $\lambda$-strongly convex in $\mathbf{w}$, adding it to the convex lifted loss establishes the second claim. 
\end{proof} 

\begin{proposition}[Degeneration to the ordinary hinge loss] \label{prop:degen_liftedloss} 
For fixed $(\mathbf{w},b,\mathbf{x})$, 
\begin{equation} \label{eq:lifted_to_hinge} 
    \lim_{\boldsymbol{\Sigma}\to0} \ell_{\Sigma}(\mathbf{w},b) = \max\left\{ 0, 1-y(\mathbf{w}^\top \mathbf{x}+b) \right\}. 
\end{equation} 
More precisely,
\begin{equation} \label{eq:lifted_hinge_error_bound} 
    \left| \ell_{\Sigma}(\mathbf{w},b) - \max\left\{ 0, 1-y(\mathbf{w}^\top \mathbf{x}+b) \right\} \right| \leq \|\mathbf{w}\|\sqrt{\operatorname{tr}(\boldsymbol{\Sigma})}. 
\end{equation} 
\end{proposition} 

\begin{proof} 
Let 
\[ 
h(z) = \max\{0,1-y(\mathbf{w}^\top \mathbf{z}+b)\}. 
\]
Since $h$ is $\|\mathbf{w}\|$-Lipschitz, 
\[ 
|h(\mathbf{x}+\xi)-h(\mathbf{x})| \leq \|\mathbf{w}\|\|\xi\|. 
\]
Therefore, 
\begin{align*} 
    |\ell_{\Sigma}(\mathbf{w},b)-h(\mathbf{x})| &\leq \mathbb{E}|h(\mathbf{x}+\xi)-h(\mathbf{x})| \\ 
    &\leq \|\mathbf{w}\|\mathbb{E}\|\xi\| \\ 
    &\leq \|\mathbf{w}\| \sqrt{\mathbb{E}\|\xi\|^2} = \|\mathbf{w}\|\sqrt{\operatorname{tr}(\boldsymbol{\Sigma})}, 
\end{align*} 
where $\xi\sim N(0,\boldsymbol{\Sigma})$. The right-hand side converges to zero as $\boldsymbol{\Sigma}\to0$, which proves both \eqref{eq:lifted_hinge_error_bound} and \eqref{eq:lifted_to_hinge}. 
\end{proof} 

The preceding result establishes convergence of the loss for fixed model parameters. The next result shows that, under standard identifiability and compactness conditions, the solutions of the lifted and ordinary SVM problems are also asymptotically equivalent. 

\begin{proposition}[Convergence of minimizers] \label{prop:lifted_minimizer_convergence} 
Consider 
\begin{equation} \label{eq:lifted_full_objective} 
    \mathcal{L}_{\Sigma}(\mathbf{w},b) = \frac{1}{2}\|\mathbf{w}\|^2 + C\sum_{i=1}^n \ell_{\Sigma}^{(i)}(\mathbf{w},b), 
\end{equation}
where 
\[ 
\ell_{\Sigma}^{(i)}(\mathbf{w},b) = \mathbb{E}_{\xi_i\sim N(0,\boldsymbol{\Sigma})} \left[ \max\left\{ 0, 1-y_i\bigl(\mathbf{w}^\top(\mathbf{x}_i+\xi_i)+b\bigr) \right\} \right], 
\] 
and let 
\begin{equation} \label{eq:ordinary_full_objective} 
    \mathcal{L}_{0}(\mathbf{w},b) = \frac{1}{2}\|\mathbf{w}\|^2 + C\sum_{i=1}^n \max\left\{ 0, 1-y_i(\mathbf{w}^\top \mathbf{x}_i+b) \right\}. 
\end{equation} 
Assume that $\mathcal{L}_0$ has a unique minimizer $(\mathbf{w}_0,b_0)$ and that the minimizers $(\mathbf{w}_{\Sigma},b_{\Sigma})$ of $\mathcal{L}_{\Sigma}$ remain in a common compact parameter set as $\boldsymbol{\Sigma}\to0$. Then 
\begin{equation} \label{eq:minimizer_convergence} 
    (\mathbf{w}_{\Sigma},b_{\Sigma}) \longrightarrow (\mathbf{w}_0,b_0), 
\end{equation} 
and 
\begin{equation} \label{eq:optimal_value_convergence} 
    \mathcal{L}_{\Sigma}(\mathbf{w}_{\Sigma},b_{\Sigma}) \rightarrow \mathcal{L}_0(\mathbf{w}_0,b_0). 
\end{equation} 
\end{proposition} 

\begin{proof} 
By Proposition~\ref{prop:degen_liftedloss}, for each observation, 
\[ 
\left| \ell_{\Sigma}^{(i)}(\mathbf{w},b) - \max\{0,1-y_i(\mathbf{w}^\top \mathbf{x}_i+b)\} \right| \leq \|\mathbf{w}\|\sqrt{\operatorname{tr}(\boldsymbol{\Sigma})}. 
\]
Hence, on any compact parameter set, 
\begin{equation}\label{eq:objective_uniform_bound} 
    |\mathcal{L}_{\Sigma}(\mathbf{w},b)-\mathcal{L}_0(\mathbf{w},b)| \leq Cn\|\mathbf{w}\|\sqrt{\operatorname{tr}(\boldsymbol{\Sigma})} \longrightarrow0 
\end{equation} 
uniformly as $\boldsymbol{\Sigma}\to0$. Thus $\mathcal{L}_{\Sigma}\to\mathcal{L}_0$ uniformly on the common compact set containing the minimizers. Since $\mathcal{L}_0$ has a unique minimizer, standard argmin consistency yields $(\mathbf{w}_{\Sigma},b_{\Sigma})\to(\mathbf{w}_0,b_0)$. Uniform convergence then also implies convergence of the corresponding minimum objective values.
\end{proof} 

Propositions~\ref{prop:lifted_smoothness}-\ref{prop:lifted_minimizer_convergence} characterize the role of the Gaussian lifting from both optimization and approximation perspectives. For nondegenerate clouds, the lifted loss is smooth and convex, while quadratic regularization provides strong convexity in $\mathbf{w}$. This permits standard gradient-based optimization of the lifted objective. At the same time, the lifted loss and, under the stated conditions, its minimizer approach their ordinary SVM counterparts as $\boldsymbol{\Sigma}\to0$. The differentiability of the nondegenerate lifted loss is especially useful for constructing representatives by matching local gradient information, which we develop next.

\subsubsection{Clouded Representatives for SVM} \label{sec:linear_rep}

Following the class structure of binary classification, representatives are constructed separately for each class on each node. Let $\mathcal{I}_{k,c}$ denote the indices of observations belonging to class $c\in\{-1,1\}$ on node $k$, with $y_i=c$, $i\in\mathcal{I}_{k,c}$, and let $n_{k,c}=|\mathcal{I}_{k,c}|$. We represent this group by the clouded pseudo-observation
\begin{align}\label{eq:clouded_rep_tuple} 
    \tilde{\mathcal D}_{k,c} = \left( n_{k,c}, \tilde y_{k,c}, \tilde{\mathbf{x}}_{k,c}, \tilde{\boldsymbol{\Sigma}}_{k,c} \right), 
\end{align}
where $\tilde y_{k,c}=c$. Thus, the representative preserves the class label and sample weight while replacing the original covariates by a representative location $\tilde{\mathbf{x}}_{k,c}$ and covariance $\tilde{\boldsymbol{\Sigma}}_{k,c}$.

For each observation $i\in\mathcal{I}_{k,c}$, define 
\begin{align*} 
    \nu_i &= c(\mathbf{w}^\top\mathbf{x}_i+b),\\ s_{k,c} &= \sqrt{\mathbf{w}^\top\boldsymbol{\Sigma}_{k,c}w},\\ a_i &= \frac{1-\nu_i}{s_{k,c}},
\end{align*} 
where $\boldsymbol{\Sigma}_{k,c}$ is the covariance used to lift observations in class $c$ on node $k$. For the clouded representative, define correspondingly 
\begin{align*}
    \tilde{\nu}_{k,c} &= c(\mathbf{w}^\top\tilde{\mathbf{x}}_{k,c}+b),\\
    \tilde{s}_{k,c} &= \sqrt{ \mathbf{w}^\top \tilde{\boldsymbol{\Sigma}}_{k,c} w },\\ 
    \tilde{a}_{k,c} &= \frac{ 1-\tilde{\nu}_{k,c} }{ \tilde{s}_{k,c} }. 
\end{align*}
We consider two constructions that differ in the information they preserve: a model-free clouded MR and a model-dependent clouded SMR.

\paragraph{Clouded Mean Representative (MR).} 
The simplest construction uses the class-wise covariate mean as the representative location, 
\begin{align} 
    \tilde{\mathbf{x}}_{k,c} = \frac{1}{n_{k,c}} \sum_{i\in\mathcal{I}_{k,c}} \mathbf{x}_i. \label{eq:clouded_mr_location}
\end{align} 
We retain the class-wise lifting covariance by setting 
\begin{align} 
    \tilde{\boldsymbol{\Sigma}}_{k,c} = \boldsymbol{\Sigma}_{k,c}. \label{eq:clouded_mr_covariance} 
\end{align} 
Together with $\tilde y_{k,c}=c$, equations \eqref{eq:clouded_mr_location}-\eqref{eq:clouded_mr_covariance} define the clouded MR 
\begin{align*} 
    \tilde{\mathcal D}_{k,c}^{\mathrm{MR}} = \left( n_{k,c}, c, \tilde{\mathbf{x}}_{k,c}, \boldsymbol{\Sigma}_{k,c} \right). 
\end{align*} 
The representative location is therefore determined entirely by the local data and does not depend on the current SVM parameters. Clouded MR can consequently be interpreted as a distributional summary of the class-wise local data rather than a local approximation constructed for a particular model iterate.

\paragraph{Clouded Score-Matching Representative (SMR).} 
Clouded MR summarizes the local data distribution but does not use the current learning objective. To incorporate optimization information, we construct a clouded Score-Matching Representative (SMR) by matching the gradient contribution of the observations in $\mathcal{I}_{k,c}$ with that of a single weighted clouded representative. In this context, ``score'' refers to the gradient of the lifted SVM objective. 

From equations \eqref{eq:lifted_gradient_w} and \eqref{eq:lifted_gradient_b}, matching the derivative with respect to $b$ gives 
\begin{align}\label{eq:smr_bias} 
    \frac{1}{n_{k,c}} \sum_{i\in\mathcal{I}_{k,c}} \Phi(a_i) = \Phi(\tilde a_{k,c}). 
\end{align} 
Since $\Phi$ is strictly increasing, 
\begin{align} 
    \tilde a_{k,c} = \Phi^{-1} \left\{ \frac{1}{n_{k,c}} \sum_{i\in\mathcal{I}_{k,c}} \Phi(a_i) \right\}. \label{eq:smr_atilde} 
\end{align} 

Matching the gradient with respect to $\mathbf{w}$ gives 
\begin{align} \label{eq:smr_weight} 
    \frac{1}{n_{k,c}} \sum_{i\in\mathcal{I}_{k,c}} \left[ -c\Phi(a_i)\mathbf{x}_i + \phi(a_i) \frac{\boldsymbol{\Sigma}_{k,c}w}{s_{k,c}} \right]  = -c\Phi(\tilde a_{k,c}) \tilde{\mathbf{x}}_{k,c} + \phi(\tilde a_{k,c}) \frac{ \tilde{\boldsymbol{\Sigma}}_{k,c}w }{ \tilde{s}_{k,c} }. 
\end{align} 
Equations \eqref{eq:smr_bias} and \eqref{eq:smr_weight} require the clouded representative to reproduce both components of the class-wise average gradient at the parameter values $(\mathbf{w},b)$ at which the representative is constructed.

To obtain a tractable covariance representation, we restrict the representative covariance to be proportional to the class-wise covariance, 
\begin{align}\label{eq:smr_covariance_form}  
    \tilde{\boldsymbol{\Sigma}}_{k,c} = \delta_{\Sigma,k,c}\boldsymbol{\Sigma}_{k,c}, 
\end{align} 
where $\delta_{\Sigma, k,c}>0$ is determined by the score-matching equations. This retains the covariance geometry of the local class while allowing its overall scale to adjust to match the lifted gradient.

Premultiplying \eqref{eq:smr_weight} by $\mathbf{w}^\top$ gives 
\begin{align}\label{eq:smr_projected} 
    \mathcal G_{k,c} = -c\Phi(\tilde a_{k,c}) \mathbf{w}^\top\tilde{\mathbf{x}}_{k,c} + \phi(\tilde a_{k,c}) \tilde{s}_{k,c}, 
\end{align} 
where
\begin{align}\label{eq:smr_G} 
    \mathcal G_{k,c} = \frac{1}{n_{k,c}} \sum_{i\in\mathcal{I}_{k,c}} \left[ -c\Phi(a_i)\mathbf{w}^\top\mathbf{x}_i + \phi(a_i)s_{k,c} \right]. 
\end{align} 
Using the definition of $\tilde a_{k,c}$, 
\begin{align}\label{eq:smr_s_relation}  
    \tilde{s}_{k,c} = \frac{ 1-c(\mathbf{w}^\top\tilde{\mathbf{x}}_{k,c}+b) }{ \tilde a_{k,c} }. 
\end{align}
Substitution into \eqref{eq:smr_projected} yields 
\begin{align}\label{eq:smr_projected_location} 
    \mathbf{w}^\top\tilde{\mathbf{x}}_{k,c} = \left[ c \left\{ \Phi(\tilde a_{k,c}) + \frac{ \phi(\tilde a_{k,c}) }{ \tilde a_{k,c} } \right\} \right]^{-1} \times \left[ (1-cb) \frac{ \phi(\tilde a_{k,c}) }{ \tilde a_{k,c} } - \mathcal G_{k,c} \right].
\end{align} 
Once the projected location is obtained, $\tilde{s}_{k,c}$ follows from \eqref{eq:smr_s_relation}. From \eqref{eq:smr_covariance_form}, 
\begin{align}\label{eq:smr_delta} 
    \delta_{\Sigma,k,c} = \frac{ \tilde{s}_{k,c}^2 }{ \mathbf{w}^\top\boldsymbol{\Sigma}_{k,c}w }, 
\end{align} 
which determines $\tilde{\boldsymbol{\Sigma}}_{k,c}$.

Finally, rearranging the full gradient-matching equation \eqref{eq:smr_weight} gives 
\begin{align} \label{eq:smr_location} 
    \tilde{\mathbf{x}}_{k,c} = \left[ c\Phi(\tilde a_{k,c}) \right]^{-1} \Bigg\{ \phi(\tilde a_{k,c}) \frac{ \tilde{\boldsymbol{\Sigma}}_{k,c}w }{ \tilde{s}_{k,c} }  - \frac{1}{n_{k,c}} \sum_{i\in\mathcal{I}_{k,c}} \left[ -c\Phi(a_i)\mathbf{x}_i + \phi(a_i) \frac{ \boldsymbol{\Sigma}_{k,c}w }{ s_{k,c} } \right] \Bigg\}. 
\end{align} 
Unlike clouded MR, clouded SMR depends on the current parameters $(\mathbf{w},b)$ and is therefore reconstructed as the model evolves. At its construction point, it preserves the class-wise gradient contribution of the local lifted SVM objective. Thus, MR provides a model-independent summary of local class location and spread, whereas SMR additionally encodes first-order information from the current lifted SVM objective.

\subsubsection{Distributed Representative Learning} \label{sec:dsvm_repl}

We now incorporate the clouded representatives into distributed SVM training. Let 
\begin{align*}
    \mathcal R_k = \{ \tilde{\mathcal D}_{h,c}: h\neq k,\; c\in\{-1,1\} \}
\end{align*}
denote the representatives received by node $k$. Node $k$ augments its local objective with the representatives received from the other nodes by minimizing the representative augmented lifted objective
\begin{align}\label{eq:distributed_repl_objective} 
    \mathcal L_k(\mathbf{w},b) = \frac{1}{2}\|\mathbf{w}\|^2 + C \sum_{i\in\mathcal I_k} \tilde{\ell}_i(\mathbf{w},b) + C \sum_{h\neq k} \sum_{c\in\{-1,1\}} n_{h,c} \tilde{\ell}_{h,c}(\mathbf{w},b), 
\end{align} 
where 
\begin{align}\label{eq:representative_lifted_loss} 
\tilde{\ell}_{h,c}(\mathbf{w},b) = (1-\tilde{\nu}_{h,c}) \Phi(\tilde a_{h,c}) + \tilde s_{h,c} \phi(\tilde a_{h,c}).
\end{align} 

The weight $n_{h,c}$ accounts for the number of observations summarized by the corresponding representative. Thus, each remote class contributes to the local optimization through a single distribution-valued pseudo-observation rather than through its individual observations.

\paragraph{Representative exchange and local updates.}
Let $\theta_k^{(t)} = \left(\mathbf{w}_k^{(t)},b_k^{(t)}\right)$ denote the parameter on node $k$ at the beginning of communication round $t$. At each outer iteration, node $k$ constructs one clouded representative for each locally observed class,
\begin{align}\label{eq:rep_round_t}
    \tilde{\mathcal D}_{k,c}^{(t)} = \left( n_{k,c}, c, \tilde{\mathbf{x}}_{k,c}^{(t)}, \tilde{\boldsymbol{\Sigma}}_{k,c}^{(t)} \right), \quad c\in\{-1,1\}.  
\end{align} 
For clouded MR, the representative is independent of the current model and therefore need not be reconstructed across communication rounds. For clouded SMR, the representative depends on the current parameter $\theta_k^{(t)}$ and is recomputed before each communication round.

The representatives are then communicated to the other participating nodes, either directly or through a coordinating server. No raw observations, gradients, or model parameters are required to construct the representative-augmented objective on another node.

After receiving the representatives from the other nodes, node $k$ forms the representative-augmented objective \eqref{eq:distributed_repl_objective}. Thus, the raw observations on node $h$ are replaced on node $k$ by their class-wise clouded representatives, with $n_{h,c}$ retaining the contribution of the number of observations summarized by each representative.

The representatives remain fixed during the local optimization phase. Starting from $\theta_{k,0}^{(t)} = \theta_k^{(t)}$,  node $k$ performs $J$ gradient steps on \eqref{eq:distributed_repl_objective}, 
\begin{align}\label{eq:local_repl_update} 
    \theta_{k,j+1}^{(t)} = \theta_{k,j}^{(t)} - \gamma_t \nabla \tilde{\mathcal L}_k^{(t)} \left( \theta_{k,j}^{(t)} \right),   
\end{align} 
for $j=0,\ldots,J-1$, where $\theta_{k,j}^{(t)}=(\mathbf{w}_{k,j}^{(t)},b_{k,j}^{(t)})$ and $\gamma_t>0$ is the local step size. At the end of the local optimization phase, 
\begin{align}\label{eq:outer_repl_update} 
\theta_k^{(t+1)} = \theta_{k,J}^{(t)}.  
\end{align}
The procedure then repeats: model-dependent representatives are reconstructed at the updated parameters, communicated, and held fixed during the next sequence of local updates.

The number of local steps $J$ controls the trade-off between local computation and communication frequency. When $J=1$, representatives are refreshed after every gradient step. Larger $J$ reduces the frequency of communication but allows the model to move farther from the parameter value at which a clouded SMR was constructed. This distinction is less consequential for clouded MR because its construction is independent of the current model.

\paragraph{Aggregation through representatives.} 
Unlike parameter-aggregation methods, the proposed procedure does not average the local parameters $\theta_k^{(t)}$ after each communication round. Instead, information from the distributed data is aggregated through the representative-augmented objectives \eqref{eq:distributed_repl_objective}. Each node retains its own parameter iterate while optimizing against its local observations and the same collection of communicated clouded representatives. Consequently, agreement between local models, when it occurs, results from their optimization against shared representative information rather than from an explicit parameter-consensus step.

\begin{algorithm}[t] 
\caption{Distributed Representative Learning for Linear SVM} \label{alg:linear_repl} 
\begin{algorithmic}[1] 
\Require Local datasets $\{\mathcal D_k\}_{k=1}^K$, number of local steps $J$, step sizes $\{\gamma_t\}$, regularization parameter $C$. 
\State Initialize $\theta_k^{(0)}=(\mathbf{w}_k^{(0)},b_k^{(0)})$ for each node $k$. 
\For{$t=0,1,\ldots,T-1$} 
    \ForAll{nodes $k$} 
        \For{$c\in\{-1,1\}$} 
            \State Construct $\tilde{\mathcal D}_{k,c}^{(t)}$ using clouded MR or clouded SMR. 
        \EndFor 
        \State Communicate $\{\tilde{\mathcal D}_{k,c}^{(t)}\}_c$ to the participating nodes. 
    \EndFor 
    \ForAll{nodes $k$} 
        \State Receive $\{\tilde{\mathcal D}_{h,c}^{(t)} :h\neq k\}$. 
        \State Set $\theta_{k,0}^{(t)}=\theta_k^{(t)}$. 
        \For{$j=0,\ldots,J-1$} 
            \State $\theta_{k,j+1}^{(t)} \gets \theta_{k,j}^{(t)} - \gamma_t \nabla \tilde{\mathcal L}_k^{(t)} (\theta_{k,j}^{(t)})$. 
        \EndFor 
        \State $\theta_k^{(t+1)} \gets \theta_{k,J}^{(t)}$. 
    \EndFor
\EndFor 
\Ensure Final local models $\{\theta_k^{(T)}\}_{k=1}^K$. 
\end{algorithmic} 
\end{algorithm}

\subsubsection{Theoretical Analysis}\label{sec:linear_theory}

Clouded SMR matches the class-wise gradient of the lifted SVM objective at the parameter value used to construct the representative. Once constructed, however, the representative remains fixed during the subsequent local optimization steps. We therefore study how its location and covariance depend on within-group heterogeneity, how well the frozen representative approximates the full-data gradient away from its construction point, and how this approximation affects the optimization iterates.

For each node $k$ and class $c\in\{-1,1\}$, define 
\begin{align*}
    \mathbf X_{k,c} &= \{ \mathbf{x}_i: i\in\mathcal I_{k,c} \},\\
    \Delta_{k,c} &= \max_{i,j\in\mathcal I_{k,c}} \| \mathbf{x}_i-\mathbf{x}_j \|,\\ 
    \Delta &= \max_{k,c}\Delta_{k,c}. 
\end{align*} 
Thus, $\Delta_{k,c}$ measures the within-group covariate diameter, while $\Delta$ is the largest diameter among the node-class groups.

We first examine a single group and suppress the indices $(k,c)$. Let $\boldsymbol{\theta}_0 = (\mathbf{w}_0,b_0)$ denote the parameter at which the clouded SMR is constructed. Define the construction-point quantities 
\begin{align*} 
    s_0 &= \sqrt{ \mathbf{w}_0^\top \boldsymbol{\Sigma} \mathbf{w}_0 },\\ 
    \nu_{i,0} &= c(\mathbf{w}_0^\top\mathbf{x}_i+b_0),\\ 
    a_{i,0} &= \frac{1-\nu_{i,0}}{s_0}. 
\end{align*} 
The representative covariance is assumed to have the proportional form 
\begin{align}
    \tilde{\boldsymbol{\Sigma}} = \delta_{\Sigma} \boldsymbol{\Sigma}, \label{eq:theory_covariance_scaling} 
\end{align}
where $\delta_{\Sigma}>0$. Define 
\begin{align*} 
    \tilde{s}_0 &= \sqrt{ \mathbf{w}_0^\top \tilde{\boldsymbol{\Sigma}} \mathbf{w}_0 },\\ \tilde{\nu}_0 &= c( \mathbf{w}_0^\top \tilde{\mathbf{x}} +b_0 ),\\ 
    \tilde{a}_0 &= \frac{ 1-\tilde{\nu}_0 }{ \tilde{s}_0 }. 
\end{align*} 
At $\boldsymbol{\theta}_0$, the score-matching conditions \eqref{eq:smr_bias} and \eqref{eq:smr_weight}, with the group indices suppressed, become 
\begin{align}
    \frac{1}{n} \sum_{i=1}^n \Phi(a_{i,0}) &= \Phi(\tilde{a}_0), \label{eq:theory_bias_matching}\\ 
    \frac{1}{n} \sum_{i=1}^n \left[ -c\Phi(a_{i,0})\mathbf{x}_i + \phi(a_{i,0}) \frac{ \boldsymbol{\Sigma}\mathbf{w}_0 }{ s_0 } \right] &= -c\Phi(\tilde{a}_0) \tilde{\mathbf{x}} + \phi(\tilde{a}_0) \frac{ \tilde{\boldsymbol{\Sigma}} \mathbf{w}_0 }{ \tilde{s}_0 }. \label{eq:theory_weight_matching} 
\end{align}

\begin{lemma}[SMR covariance-scale stability] \label{lem:smr_covariance_scale} 
Let 
\begin{align*} 
    \mathcal U_r^{(0)} = \left\{ (\mathbf{w},b): \| (\mathbf{w},b) - (\mathbf{w}_0,b_0) \| \leq r \right\}. 
\end{align*}
Suppose that $\boldsymbol{\Sigma}$ is symmetric and positive definite and that there exist constants $M_w,M_b,M_x,\epsilon_s>0$ such that, for all $(\mathbf{w},b)\in\mathcal U_r^{(0)}$, 
\begin{align*} 
    \|\mathbf{w}\| &\leq M_w,\\ 
    |b| &\leq M_b,\\
    \max_i\|\mathbf{x}_i\| &\leq M_x,\\ 
    \sqrt{ \mathbf{w}^\top \boldsymbol{\Sigma} \mathbf{w} } &\geq\epsilon_s. 
\end{align*}
Define 
\begin{align*}
    A_r &= \frac{ 1+M_wM_x+M_b }{ \epsilon_s },\\ 
    \mathcal A_r &= [-A_r,A_r],\\ 
    m_{\phi,r} &= \inf_{a\in\mathcal A_r}\phi(a),\\
    h_{\min,r} &= \inf_{a\in\mathcal A_r}h(a), 
\end{align*} 
where $h(a) = a\Phi(a)+\phi(a)$.  Then $m_{\phi,r}>0$, $h_{\min,r}>0$, and 
\begin{align}\label{eq:covariance_scale_identity} 
    \sqrt{\delta_{\Sigma}} = \frac{ n^{-1}\sum_{i=1}^n h(a_{i,0}) }{ h(\tilde{a}_0) }. 
\end{align}
Moreover, 
\begin{align}\label{eq:covariance_scale_bound}
    \left| \sqrt{\delta_{\Sigma}}-1 \right| \leq \frac{M_w}{ h_{\min,r}\epsilon_s } \left( 1+ \frac{L_\Phi}{m_{\phi,r}} \right) \Delta,  
\end{align} 
where $L_\Phi = \sup_{a\in\mathbb R}\phi(a) = \frac{1}{\sqrt{2\pi}}$. 
Consequently, $\left| \sqrt{\delta_{\Sigma}}-1 \right| = \mathcal O(\Delta)$. 
For $\Delta$ sufficiently small,  $|\delta_{\Sigma}-1| = \mathcal O(\Delta)$. 
\end{lemma}

\begin{proof} 
From \eqref{eq:theory_covariance_scaling}, 
\begin{align*} 
    \tilde{s}_0 = \sqrt{ \mathbf{w}_0^\top \delta_{\Sigma} \boldsymbol{\Sigma} \mathbf{w}_0 }= \sqrt{\delta_{\Sigma}}\,s_0. 
\end{align*} 
Premultiplying \eqref{eq:theory_weight_matching} by $\mathbf{w}_0^\top$ gives 
\begin{align}\label{eq:theory_projected_matching}
    -c\Phi(\tilde a_0) \mathbf{w}_0^\top \tilde{\mathbf{x}} + \phi(\tilde a_0) \tilde s_0 &= \frac{1}{n} \sum_{i=1}^n \left[ -c\Phi(a_{i,0}) \mathbf{w}_0^\top \mathbf{x}_i + \phi(a_{i,0})s_0 \right].  
\end{align} 
From the definitions of the standardized margins, 
\begin{align*} 
    c\mathbf{w}_0^\top \tilde{\mathbf{x}} &= 1-cb_0- \tilde a_0\tilde s_0,\\
    c\mathbf{w}_0^\top \mathbf{x}_i &= 1-cb_0-a_{i,0}s_0. 
\end{align*} 
Substituting these expressions into \eqref{eq:theory_projected_matching} and applying \eqref{eq:theory_bias_matching} cancels the terms containing $1-cb_0$. We obtain 
\begin{align*}
    h(\tilde a_0) \tilde s_0 = s_0 \frac{1}{n} \sum_{i=1}^n h(a_{i,0}). 
\end{align*}
Since $\tilde s_0/s_0=\sqrt{\delta_{\Sigma}}$, this proves \eqref{eq:covariance_scale_identity}.

The assumed parameter bounds imply 
\begin{align*} 
    |a_{i,0}| = \frac{ | 1-c( \mathbf{w}_0^\top \mathbf{x}_i+b_0 ) | }{ s_0 }\leq A_r.
\end{align*} 
Therefore, $a_{i,0}\in\mathcal A_r$. By \eqref{eq:theory_bias_matching} and monotonicity of $\Phi$, $\tilde a_0\in\mathcal A_r$ as well. Hence $m_{\phi,r}>0$. Also,  $h'(a)=\Phi(a)$, and $h(a)>0$ for every finite $a$, so continuity on the compact interval $\mathcal A_r$ gives $h_{\min,r}>0$.

Let 
\begin{align*} 
\overline a_0 &= \frac{1}{n} \sum_{i=1}^n a_{i,0},\\ 
\overline h_0 &= \frac{1}{n} \sum_{i=1}^n h(a_{i,0}). 
\end{align*} 
For any $i,j$, 
\begin{align*} 
    |a_{i,0}-a_{j,0}| = \frac{ | \mathbf{w}_0^\top (\mathbf{x}_i-\mathbf{x}_j) | }{ s_0 }\leq \frac{M_w}{\epsilon_s}\Delta. 
\end{align*} 
It follows that 
\begin{align}\label{eq:ai_mean_bound} 
    |a_{i,0}-\overline a_0| \leq \frac{M_w}{\epsilon_s}\Delta. 
\end{align} 
Because $h'(a)=\Phi(a)$ and $0\leq\Phi(a)\leq1$, $h$ is $1$-Lipschitz. Therefore, 
\begin{align} 
    | \overline h_0-h(\overline a_0) | &\leq \frac{1}{n} \sum_{i=1}^n | a_{i,0}-\overline a_0 | \nonumber\\
    &\leq \frac{M_w}{\epsilon_s}\Delta. \label{eq:h_average_bound}
\end{align}

Next, by \eqref{eq:theory_bias_matching}, 
\begin{align*} 
    \Phi(\tilde a_0) = \frac{1}{n} \sum_{i=1}^n \Phi(a_{i,0}). 
\end{align*} 
Using the $L_\Phi$-Lipschitz property of $\Phi$ and \eqref{eq:ai_mean_bound}, 
\begin{align*} 
    | \Phi(\tilde a_0) - \Phi(\overline a_0) | &\leq \frac{L_\Phi}{n} \sum_{i=1}^n | a_{i,0}-\overline a_0 |\\
    &\leq L_\Phi \frac{M_w}{\epsilon_s}\Delta.
\end{align*} 
Since both arguments lie in $\mathcal A_r$, the mean value theorem and the definition of $m_{\phi,r}$ give 
\begin{align}\label{eq:atilde_mean_bound} 
    | \tilde a_0-\overline a_0 | \leq \frac{L_\Phi}{m_{\phi,r}} \frac{M_w}{\epsilon_s}\Delta. 
\end{align}
Using again the $1$-Lipschitz property of $h$,
\begin{align}\label{eq:h_tilde_mean_bound} 
    | h(\tilde a_0) - h(\overline a_0) | \leq \frac{L_\Phi}{m_{\phi,r}} \frac{M_w}{\epsilon_s}\Delta. 
\end{align}

By \eqref{eq:covariance_scale_identity}, 
\begin{align*} 
    \left| \sqrt{\delta_{\Sigma}}-1 \right| &= \frac{ | \overline h_0 - h(\tilde a_0) | }{ h(\tilde a_0) }. 
\end{align*} 
Combining \eqref{eq:h_average_bound} and \eqref{eq:h_tilde_mean_bound}, and using $h(\tilde a_0)\geq h_{\min,r}$, proves \eqref{eq:covariance_scale_bound}.

Finally, 
\begin{align*} 
    |\delta_{\Sigma}-1| = \left| \sqrt{\delta_{\Sigma}}-1 \right| \left( \sqrt{\delta_{\Sigma}}+1 \right). 
\end{align*} 
For sufficiently small $\Delta$, $\sqrt{\delta_{\Sigma}}$ remains bounded, and hence $|\delta_{\Sigma}-1|=\mathcal O(\Delta)$. 
\end{proof}

Lemma~\ref{lem:smr_covariance_scale} is a local stability result rather than a degenerating-covariance asymptotic result. For an empirical within-group covariance, $\|\widehat{\boldsymbol{\Sigma}}\|$ may itself decrease with the group diameter. In implementation, we stabilize the covariance through $\boldsymbol{\Sigma} = \widehat{\boldsymbol{\Sigma}} + \eta I$,  where $\eta>0$, so that the lifted standard deviation remains bounded away from zero.

We next compare the gradient evaluated on the full distributed data with the gradient evaluated on fixed clouded SMR representatives.

\begin{theorem}[Gradient discrepancy for frozen clouded SMR] \label{thm:smr_gradient_discrepancy} 
At a construction point $\boldsymbol{\theta}_{0,k} = (\mathbf{w}_{0,k},b_{0,k})$, construct the clouded representative $\tilde{\mathbf X}_{k,c} = \left( \tilde{\mathbf{x}}_{k,c}, \tilde{\boldsymbol{\Sigma}}_{k,c} \right)$ satisfying the bias- and weight-matching equations, with  $\tilde{\boldsymbol{\Sigma}}_{k,c} = \delta_{\Sigma,k,c} \boldsymbol{\Sigma}_{k,c}$. 

Suppose the assumptions of Lemma~\ref{lem:smr_covariance_scale} hold uniformly over all groups. In particular, suppose 
\begin{align}\label{eq:uniform_covariance_scale} 
    \left| \sqrt{\delta_{\Sigma,k,c}}-1 \right| \leq C_\Sigma \Delta_{k,c} 
\end{align} 
for a constant $C_\Sigma$ independent of $(k,c)$.

For $r>0$, define 
\begin{align*} 
    \mathcal U_{k,r} = \left\{ \boldsymbol{\theta}: \| \boldsymbol{\theta} - \boldsymbol{\theta}_{0,k} \| \leq r \right\}, 
\end{align*} 
and suppose $\mathcal U_r = \bigcap_k\mathcal U_{k,r}$ is nonempty. Assume uniformly over all groups and all $\boldsymbol{\theta}=(\mathbf{w},b)\in\mathcal U_{k,r}$ that 
\begin{align*} 
    \|\mathbf{w}\| &\leq M_w,\\ 
    |b| &\leq M_b,\\
    \|\mathbf{x}_i\| &\leq M_x,\\
    \| \boldsymbol{\Sigma}_{k,c} \| &\leq M_\Sigma,\\
    s_{k,c}(\mathbf{w}) &= \sqrt{ \mathbf{w}^\top \boldsymbol{\Sigma}_{k,c} \mathbf{w} } \geq\epsilon_s,\\
    \tilde s_{k,c}(\mathbf{w}) &= \sqrt{ \mathbf{w}^\top \tilde{\boldsymbol{\Sigma}}_{k,c} \mathbf{w} } \geq\epsilon_s. 
\end{align*} 
Assume also that the original and representative standardized margins remain in a common compact interval.

Define the full-data and frozen-representative objectives by 
\begin{align}
    \mathcal L(\boldsymbol{\theta}) &= \frac{1}{2}\|\mathbf{w}\|^2 + C \sum_{k,c} \alpha_{k,c} \ell_c( \mathbf X_{k,c}; \boldsymbol{\theta} ), \label{eq:theory_full_objective}\\ 
    \tilde{\mathcal L}(\boldsymbol{\theta}) &= \frac{1}{2}\|\mathbf{w}\|^2 + C \sum_{k,c} \alpha_{k,c} \ell_c( \tilde{\mathbf X}_{k,c}; \boldsymbol{\theta} ), \label{eq:theory_representative_objective} 
\end{align} 
where $\alpha_{k,c} \geq0$, $\sum_{k,c}\alpha_{k,c} =1$. 
Then there exists a constant $\kappa >0$, independent of $\Delta$, such that 
\begin{align}\label{eq:global_gradient_discrepancy} 
    \sup_{\boldsymbol{\theta}\in\mathcal U_r} \left\| \nabla\mathcal L(\boldsymbol{\theta}) - \nabla\tilde{\mathcal L} (\boldsymbol{\theta}) \right\| \leq \kappa \Delta. 
\end{align} 
\end{theorem}

\begin{proof} 
Fix a group $(k,c)$ and suppress its indices. Let 
\begin{align*} \overline{\mathbf{x}} = \frac{1}{n} \sum_{i=1}^n\mathbf{x}_i. \end{align*} We first obtain structural bounds on the representative at the construction point. Combining the weight-matching and bias-matching equations gives 
\begin{align} 
    \Phi(\tilde a_0) \left( \tilde{\mathbf{x}} - \overline{\mathbf{x}} \right) &= \frac{1}{n} \sum_{i=1}^n \Phi(a_{i,0}) \left( \mathbf{x}_i-\overline{\mathbf{x}} \right) \nonumber\\ 
    &\quad+ c \left[ \phi(\tilde a_0) \frac{ \tilde{\boldsymbol{\Sigma}} \mathbf{w}_0 }{ \tilde s_0 } - \left\{ \frac{1}{n} \sum_{i=1}^n \phi(a_{i,0}) \right\} \frac{ \boldsymbol{\Sigma}\mathbf{w}_0 }{ s_0 } \right]. \label{eq:representative_location_identity} 
\end{align} 
Because $\overline{\mathbf{x}}$ belongs to the convex hull of the group, $\| \mathbf{x}_i-\overline{\mathbf{x}} \| \leq \Delta$.  
Therefore, 
\begin{align}\label{eq:location_average_bound} 
    \left\| \frac{1}{n} \sum_{i=1}^n \Phi(a_{i,0}) ( \mathbf{x}_i-\overline{\mathbf{x}} ) \right\| \leq \Delta. 
\end{align}

Since $\tilde{\boldsymbol{\Sigma}} =\delta_{\Sigma}\boldsymbol{\Sigma}$, 
\begin{align*} 
    \frac{ \tilde{\boldsymbol{\Sigma}} \mathbf{w}_0 }{ \tilde s_0 } = \sqrt{\delta_{\Sigma}} \frac{ \boldsymbol{\Sigma}\mathbf{w}_0 }{ s_0 }. 
\end{align*} 
Lemma~\ref{lem:smr_covariance_scale}, together with the Lipschitz continuity of $\phi$ on the compact standardized-margin interval, implies 
\begin{align*}
    \left| \phi(\tilde a_0) \sqrt{\delta_{\Sigma}} - \frac{1}{n} \sum_{i=1}^n\phi(a_{i,0}) \right| \leq C_1 \Delta 
\end{align*} 
for a constant $C_1>0$. Moreover, 
\begin{align*} 
    \left\| \frac{ \boldsymbol{\Sigma}\mathbf{w}_0 }{ s_0 } \right\| \leq \frac{M_\Sigma M_w}{\epsilon_s}. 
\end{align*} 
Since $\tilde a_0$ remains in a compact interval, $\Phi(\tilde a_0)$ is bounded below by a positive constant. Applying these bounds to \eqref{eq:representative_location_identity} yields 
\begin{align}\label{eq:representative_location_bound}  
    \| \tilde{\mathbf{x}} - \overline{\mathbf{x}} \| \leq C_x \Delta 
\end{align} 
for some $C_x>0$. Consequently, 
\begin{align}\label{eq:data_representative_location_bound} 
    \| \mathbf{x}_i - \tilde{\mathbf{x}} \| \leq (1+C_x) \Delta.  
\end{align}

Now fix an arbitrary $\boldsymbol{\theta}=(\mathbf{w},b)\in\mathcal U_{k,r}$. The representative remains fixed, so the matching equations are not assumed to hold at this new parameter.

By \eqref{eq:uniform_covariance_scale}, 
\begin{align}\label{eq:scale_difference_bound} 
    |s-\tilde s| = \left| 1-\sqrt{\delta_{\Sigma}} \right|s \leq C_s \Delta 
\end{align}
for some $C_s>0$. Similarly, \eqref{eq:data_representative_location_bound} gives 
\begin{align}\label{eq:margin_difference_bound}
    |\nu_i-\tilde\nu| \leq \|\mathbf{w}\| \| \mathbf{x}_i - \tilde{\mathbf{x}} \| \leq C_\nu \Delta.  
\end{align} 
Because the model parameters and representative location are bounded, $|1-\tilde\nu|$ is uniformly bounded. Since $s,\tilde s\geq\epsilon_s$,
\begin{align*} 
    |a_i-\tilde a| &\leq \frac{ |\nu_i-\tilde\nu| }{ s } + |1-\tilde\nu| \frac{ |s-\tilde s| }{ s\tilde s }\\
    &\leq C_a\Delta 
\end{align*} for some $C_a>0$.

The bias derivatives satisfy 
\begin{align*} 
    \frac{\partial}{ \partial b } \ell_c( \mathbf X; \boldsymbol{\theta} ) &= -\frac{c}{n} \sum_{i=1}^n\Phi(a_i),\\
    \frac{\partial}{ \partial b } \ell_c( \tilde{\mathbf X}; \boldsymbol{\theta} ) &= -c\Phi(\tilde a). 
\end{align*} 
Since $\Phi$ is globally Lipschitz, 
\begin{align}\label{eq:bias_gradient_group_bound} 
    \left| \frac{\partial}{ \partial b } \ell_c( \mathbf X; \boldsymbol{\theta} ) - \frac{\partial}{ \partial b } \ell_c( \tilde{\mathbf X}; \boldsymbol{\theta} ) \right| \leq C_b \Delta. 
\end{align}

For the weight-gradient location component, 
\begin{align*} 
    & \left\| \Phi(a_i)\mathbf{x}_i - \Phi(\tilde a) \tilde{\mathbf{x}} \right\|\\
    \leq & | \Phi(a_i)-\Phi(\tilde a) | \|\mathbf{x}_i\| + \Phi(\tilde a) \| \mathbf{x}_i - \tilde{\mathbf{x}} \|\\
    \leq & C_{\mathrm{loc}}\Delta. 
\end{align*} 
Averaging over $i$ gives the same order.

For the covariance component, add and subtract $\phi(\tilde a)\boldsymbol{\Sigma}\mathbf{w}/s$. The Lipschitz continuity of $\phi$ on the compact standardized-margin interval gives 
\begin{align*} 
    \left| \frac{1}{n} \sum_{i=1}^n\phi(a_i) - \phi(\tilde a) \right| \leq C_\phi \Delta. 
\end{align*} 
Furthermore, 
\begin{align*}
    \frac{ \tilde{\boldsymbol{\Sigma}}\mathbf{w} }{ \tilde s } = \sqrt{\delta_{\Sigma}} \frac{ \boldsymbol{\Sigma}\mathbf{w} }{ s }, 
\end{align*} 
and hence
\begin{align*}
    \left\| \frac{ \tilde{\boldsymbol{\Sigma}}\mathbf{w} }{ \tilde s } - \frac{ \boldsymbol{\Sigma}\mathbf{w} }{ s } \right\| \leq C_{\mathrm{cov}}\Delta. 
\end{align*} 
Combining the location and covariance terms yields 
\begin{align}\label{eq:weight_gradient_group_bound} 
    \left\| \nabla_{\mathbf{w}} \ell_c( \mathbf X; \boldsymbol{\theta} ) - \nabla_{\mathbf{w}} \ell_c( \tilde{\mathbf X}; \boldsymbol{\theta} ) \right\| \leq C_w \Delta. 
\end{align} 
Together with \eqref{eq:bias_gradient_group_bound}, this implies 
\begin{align}\label{eq:group_gradient_bound} 
    \sup_{\boldsymbol{\theta}\in\mathcal U_{k,r}} \left\| \nabla \ell_c( \mathbf X_{k,c}; \boldsymbol{\theta} ) - \nabla \ell_c( \tilde{\mathbf X}_{k,c}; \boldsymbol{\theta} ) \right\| \leq C_g\Delta_{k,c}. 
\end{align}

Finally, the regularization gradients in \eqref{eq:theory_full_objective} and \eqref{eq:theory_representative_objective} are identical and cancel. Therefore, 
\begin{align*}
    & \left\| \nabla\mathcal L(\boldsymbol{\theta}) - \nabla\tilde{\mathcal L} (\boldsymbol{\theta}) \right\|\\
    \leq &C \sum_{k,c} \alpha_{k,c} \left\| \nabla\ell_c( \mathbf X_{k,c}; \boldsymbol{\theta} ) - \nabla\ell_c( \tilde{\mathbf X}_{k,c}; \boldsymbol{\theta} ) \right\|\\
    \leq &CC_g \sum_{k,c} \alpha_{k,c}\Delta_{k,c}\\ 
    \leq &CC_g\Delta. 
\end{align*} 
Taking the supremum over $\mathcal U_r$ and setting $\kappa=CC_g$ proves \eqref{eq:global_gradient_discrepancy}. 

\end{proof}

Theorem~\ref{thm:smr_gradient_discrepancy} describes the approximation quality of a frozen clouded SMR. Exact gradient matching holds at the construction point, while the discrepancy over a nearby optimization region is controlled to first order by the largest within-group diameter. Thus, class-wise subgroups with compact covariate geometry produce more faithful gradient representations.

We conclude by translating the gradient-discrepancy bound into an optimization guarantee for the iterative representative-learning procedure.

\begin{corollary}[Inexact gradient descent with iterative clouded SMR] \label{cor:iterative_smr_convergence} 
Let $\mathcal L:\mathbb R^{p+1}\rightarrow\mathbb R$ denote the full-data objective, and let $\boldsymbol{\theta}^*$ be its minimizer. Suppose that $\mathcal L$ is $\mu$-strongly convex and $L$-smooth on a convex neighborhood $\mathcal U$.

At outer iteration $t$, construct clouded SMR representatives at $\boldsymbol{\theta}^{(t)}$, producing the frozen representative objective $\tilde{\mathcal L}^{(t)}$. Set $\boldsymbol{\theta}_{0}^{(t)} = \boldsymbol{\theta}^{(t)}$  and perform $J$ inner updates 
\begin{align}\label{eq:theory_inner_update} 
    \boldsymbol{\theta}_{j+1}^{(t)} = \boldsymbol{\theta}_{j}^{(t)} - \gamma \nabla \tilde{\mathcal L}^{(t)} \left( \boldsymbol{\theta}_{j}^{(t)} \right),  
\end{align}
where $j=0,\ldots,J-1$. Define $\boldsymbol{\theta}^{(t+1)} = \boldsymbol{\theta}_{J}^{(t)}$.  Assume that all outer and inner iterates remain in $\mathcal U$, and that 
\begin{align}\label{eq:iterative_gradient_error}  
    \sup_{\boldsymbol{\theta}\in\mathcal U} \left\| \nabla \tilde{\mathcal L}^{(t)} (\boldsymbol{\theta}) - \nabla \mathcal L(\boldsymbol{\theta}) \right\| \leq  \kappa \Delta 
\end{align} 
uniformly over $t$. If  $0<\gamma\leq\frac{1}{L}$,  and  $\rho = 1-\gamma\mu$,  then after $T$ outer iterations, 
\begin{align}\label{eq:iterative_smr_convergence}
    \| \boldsymbol{\theta}^{(T)} - \boldsymbol{\theta}^* \| \leq \rho^{JT} \| \boldsymbol{\theta}^{(0)} - \boldsymbol{\theta}^* \| + \frac{\kappa \Delta}{\mu} \left( 1-\rho^{JT} \right).  
\end{align}
Consequently, 
\begin{align}\label{eq:iterative_smr_error_floor} 
    \limsup_{T\rightarrow\infty} \| \boldsymbol{\theta}^{(T)} - \boldsymbol{\theta}^* \| \leq \frac{\kappa \Delta}{\mu}.  
\end{align} 

\end{corollary}

\begin{proof} 
For each outer iteration $t$ and inner iteration $j$, define
\begin{align*} 
    \mathbf{e}_{t,j} = \nabla \tilde{\mathcal L}^{(t)} \left( \boldsymbol{\theta}_{j}^{(t)} \right) - \nabla \mathcal L \left( \boldsymbol{\theta}_{j}^{(t)} \right). 
\end{align*}
The update in \eqref{eq:theory_inner_update} can be written as
\begin{align*} 
    \boldsymbol{\theta}_{j+1}^{(t)} = \boldsymbol{\theta}_{j}^{(t)} - \gamma \nabla \mathcal L \left( \boldsymbol{\theta}_{j}^{(t)} \right) - \gamma\mathbf{e}_{t,j}. 
\end{align*} 
Since $\mathcal L$ is $\mu$-strongly convex and $L$-smooth, and $0<\gamma\leq1/L$, the exact gradient map is contractive: 
\begin{align*} 
    & \left\| \boldsymbol{\theta}_{j}^{(t)} - \gamma \nabla\mathcal L \left( \boldsymbol{\theta}_{j}^{(t)} \right) - \boldsymbol{\theta}^* \right\|\\ 
    \leq &  \rho \left\| \boldsymbol{\theta}_{j}^{(t)} - \boldsymbol{\theta}^* \right\|. 
\end{align*} 
Therefore, 
\begin{align*}
    \left\| \boldsymbol{\theta}_{j+1}^{(t)} - \boldsymbol{\theta}^* \right\| &\leq \rho \left\| \boldsymbol{\theta}_{j}^{(t)} - \boldsymbol{\theta}^* \right\| + \gamma \|\mathbf{e}_{t,j}\|\\ 
    &\leq \rho \left\| \boldsymbol{\theta}_{j}^{(t)} - \boldsymbol{\theta}^* \right\| + \gamma \kappa\Delta.
\end{align*} 
Iterating over the $J$ inner updates gives 
\begin{align*} 
    \left\| \boldsymbol{\theta}^{(t+1)} - \boldsymbol{\theta}^* \right\| &\leq \rho^J \left\| \boldsymbol{\theta}^{(t)} - \boldsymbol{\theta}^* \right\| + \gamma \kappa\Delta \sum_{j=0}^{J-1}\rho^j\\ 
    &= \rho^J \left\| \boldsymbol{\theta}^{(t)} - \boldsymbol{\theta}^* \right\| + \frac{\kappa\Delta}{\mu} \left( 1-\rho^J \right). 
\end{align*} 
Applying this recursion over $T$ outer iterations proves \eqref{eq:iterative_smr_convergence}. Taking the limit superior gives \eqref{eq:iterative_smr_error_floor}. 

\end{proof}

The strong-convexity assumption in Corollary~\ref{cor:iterative_smr_convergence} applies to the full parameter $\boldsymbol{\theta}=(\mathbf{w},b)$. The standard SVM regularizer guarantees strong convexity in $\mathbf{w}$ but does not regularize the intercept $b$. The corollary therefore applies directly when the intercept is fixed, eliminated, or regularized. Without such a condition, the corresponding contraction guarantee should be interpreted with respect to the regularized weight parameter.

\begin{remark}[Exact first step after reconstruction] \label{rem:exact_first_step} 
Suppose that all clouded SMR representatives are constructed at the same parameter $\boldsymbol{\theta}^{(t)}$. By score matching, 
\begin{align*} 
    \nabla \widetilde{\mathcal L}^{(t)} \left( \boldsymbol{\theta}^{(t)} \right) = \nabla \mathcal L \left( \boldsymbol{\theta}^{(t)} \right). 
\end{align*} 
Hence the first inner step after each representative reconstruction has zero gradient-approximation error. Repeating the proof of Corollary~\ref{cor:iterative_smr_convergence} over the remaining $J-1$ potentially inexact steps gives 
\begin{align*} 
    \left\| \boldsymbol{\theta}^{(t+1)} - \boldsymbol{\theta}^* \right\| \leq \rho^J \left\| \boldsymbol{\theta}^{(t)} - \boldsymbol{\theta}^* \right\| + \frac{\kappa\Delta}{\mu} \left( 1-\rho^{J-1} \right). 
\end{align*}  
In particular, when $J=1$, the representative objective produces the same gradient update as the full-data objective at every outer iteration. 
\end{remark}

\subsection{Kernel SVM}

The distributed problem becomes more involved for nonlinear SVMs. Given a feature map $\psi:\mathbb{R}^p\rightarrow\mathcal H$ into a reproducing kernel Hilbert space (RKHS) $\mathcal H$, a kernel $K(\mathbf{x},\mathbf{x}')$ satisfies 
\begin{align*} 
    K(\mathbf{x},\mathbf{x}') = \left\langle \psi(\mathbf{x}), \psi(\mathbf{x}') \right\rangle_{\mathcal H}. 
\end{align*} 
The representer theorem yields a finite expansion of the learned function in terms of kernel evaluations at observed data points. Consequently, nonlinear SVM training introduces additional storage and computational costs associated with the kernel representation. Low-dimensional or low-rank approximations, including random Fourier features and Nystr\"om approximations, provide one route for reducing these costs and can facilitate distributed implementations.

For the lifted formulation, an additional issue arises from the variance term. In the linear setting, Gaussian lifting leads to $s^2=\mathbf{w}^{\top}\boldsymbol{\Sigma}\mathbf{w}$,  which is the variance of the linear score induced by the cloud distribution. To obtain the corresponding nonlinear formulation, let 
\begin{align*} 
f(\mathbf{x}) = \left\langle w,\psi(\mathbf{x}) \right\rangle_{\mathcal H}+b, 
\end{align*} 
and let $\Gamma$ denote the covariance operator associated with $\psi(X)$, 
\begin{align}\label{eq:kernel_covariance_operator} 
\Gamma = \mathbb{E}\left[ \left(\psi(X)-\mu_\psi\right) \otimes \left(\psi(X)-\mu_\psi\right) \right], 
\end{align} 
where $\mu_\psi=\mathbb{E}[\psi(X)]$. Then 
\begin{align}\label{eq:kernel_score_variance} 
    \operatorname{Var}\{f(X)\} = \left\langle w,\Gamma w \right\rangle_{\mathcal H}. 
\end{align} 
Thus, $\langle w,\Gamma w\rangle_{\mathcal H}$ plays the role of $\mathbf{w}^{\top}\boldsymbol{\Sigma}\mathbf{w}$ in the linear lifted SVM.

Directly communicating an empirical covariance operator is generally unsuitable for distributed representative learning because the local RKHS representations may be expressed in terms of different local observations. We therefore introduce a shared dictionary  $\mathcal B = \{\boldsymbol{\xi}_1,\ldots,\boldsymbol{\xi}_m\}$,  available to all participating nodes. The dictionary serves both as a finite-dimensional model space and as a common coordinate system in which clouded representatives can be communicated.

Define the dictionary Gram matrix $\mathbf K\in\mathbb R^{m\times m}$ and the dictionary kernel vector $\mathbf{k}(\mathbf{x})\in\mathbb R^m$ by 
\begin{align}
    [\mathbf K]_{jl} &= K(\boldsymbol{\xi}_j,\boldsymbol{\xi}_l), \label{eq:kernel_dictionary_gram}\\ 
    \mathbf{k}(\mathbf{x}) &= \left( K(\boldsymbol{\xi}_1,\mathbf{x}), \ldots, K(\boldsymbol{\xi}_m,\mathbf{x}) \right)^\top. \label{eq:kernel_dictionary_vector} 
\end{align} 
We restrict the classifier to the dictionary span,
\begin{align}\label{eq:kernel_dictionary_model} 
    w = \sum_{j=1}^{m} \beta_j\psi(\boldsymbol{\xi}_j), 
\end{align} 
so that 
\begin{align*}
    f(\mathbf{x}) = \boldsymbol{\beta}^{\top}\mathbf{k}(\mathbf{x})+b, \quad \|w\|_{\mathcal H}^2 = \boldsymbol{\beta}^{\top}\mathbf K\boldsymbol{\beta}. 
\end{align*}

For a class-wise group $\mathcal D_{k,c}=\{\mathbf{x}_i:i\in\mathcal I_{k,c}\}$, let 
\begin{align*} 
    \mathbf K_{k,c} = \left[ \mathbf{k}(\mathbf{x}_i) \right]_{i\in\mathcal I_{k,c}} \in \mathbb R^{m\times n_{k,c}}, 
\end{align*} 
and let
\begin{align*}
    \mathbf H_{k,c} = \mathbf I_{n_{k,c}} - \frac{1}{n_{k,c}}\mathbf 1\mathbf 1^\top. 
\end{align*} 
We define the dictionary-based covariance matrix 
\begin{align}\label{eq:kernel_group_covariance}  
    \mathbf C_{k,c} = \frac{1}{n_{k,c}} \mathbf K_{k,c} \mathbf H_{k,c} \mathbf K_{k,c}^{\top}. 
\end{align} 
For a parameter $\boldsymbol{\beta}$, the corresponding lifted scale is 
\begin{align} 
    s_{k,c} = \sqrt{ \boldsymbol{\beta}^{\top} \mathbf C_{k,c} \boldsymbol{\beta} }. \label{eq:kernel_lifted_scale} 
\end{align}
Equivalently, 
\begin{align*}
    s_{k,c}^2 = \frac{1}{n_{k,c}} \left( \mathbf K_{k,c}^{\top}\boldsymbol{\beta} \right)^\top \mathbf H_{k,c} \left( \mathbf K_{k,c}^{\top}\boldsymbol{\beta} \right), 
\end{align*} 
so $s_{k,c}^2$ is the empirical variance of the kernel scores within the node-class group. This provides a finite-dimensional analogue of \eqref{eq:kernel_score_variance} that is computable using kernel evaluations.

For $i\in\mathcal I_{k,c}$, define 
\begin{align*} 
    \nu_i &= c\left( \boldsymbol{\beta}^{\top}\mathbf{k}(\mathbf{x}_i)+b \right),\\ a_i &= \frac{1-\nu_i}{s_{k,c}}. 
\end{align*} 
The kernel lifted hinge loss retains the same form as in the linear case, 
\begin{align}\label{eq:kernel_lifted_hinge}  
    \ell_i(\boldsymbol{\beta},b) = (1-\nu_i)\Phi(a_i) + s_{k,c}\phi(a_i), 
\end{align} 
with regularization 
\begin{align}\label{eq:kernel_regularization}
    \frac{1}{2}\|w\|_{\mathcal H}^2 = \frac{1}{2} \boldsymbol{\beta}^{\top} \mathbf K \boldsymbol{\beta}.  
\end{align} 
Hence the lifted SVM framework extends to the nonlinear setting while remaining finite-dimensional with respect to the shared dictionary.

\subsubsection{Clouded Representatives for Kernel SVM}

We next construct class-wise clouded representatives in the dictionary coordinate system. For a group $(k,c)$, let a representative be summarized by 
\begin{align} 
\tilde{\mathcal D}_{k,c} = \left( n_{k,c}, c, \boldsymbol{\gamma}_{k,c}, \tilde{\mathbf C}_{k,c} \right), \label{eq:kernel_rep_tuple} 
\end{align} 
where $\boldsymbol{\gamma}_{k,c}\in\mathbb R^m$ determines a representative in the dictionary span, 
\begin{align*} 
    \tilde z_{k,c} = \sum_{j=1}^{m} \gamma_{k,c,j}\psi(\boldsymbol{\xi}_j). 
\end{align*}
Its dictionary interaction vector is 
\begin{align}\label{eq:kernel_rep_interaction} 
    \tilde{\mathbf r}_{k,c} = \mathbf K\boldsymbol{\gamma}_{k,c}, 
\end{align} 
so that 
\begin{align*}
    \left\langle w,\tilde z_{k,c} \right\rangle_{\mathcal H} = \boldsymbol{\beta}^{\top} \tilde{\mathbf r}_{k,c}. 
\end{align*}
The representative margin and lifted scale are therefore 
\begin{align*}
    \tilde\nu_{k,c} &= c\left( \boldsymbol{\beta}^{\top}\tilde{\mathbf r}_{k,c}+b \right),\\ \tilde s_{k,c} &= \sqrt{ \boldsymbol{\beta}^{\top} \tilde{\mathbf C}_{k,c} \boldsymbol{\beta} },\\ 
    \tilde a_{k,c} &= \frac{1-\tilde\nu_{k,c}}{\tilde s_{k,c}}. 
\end{align*}

\paragraph{Kernel Mean Representative.}

The analogue of the class-wise mean representative is the empirical kernel mean. Its dictionary interaction vector is 
\begin{align}\label{eq:kernel_mr_interaction}
    \mathbf r_{k,c}^{\mathrm{MR}} = \frac{1}{n_{k,c}} \sum_{i\in\mathcal I_{k,c}} \mathbf{k}(\mathbf{x}_i).  
\end{align} 
A dictionary-span realization is obtained by choosing $\boldsymbol{\gamma}_{k,c}^{\mathrm{MR}}$ to satisfy 
\begin{align}\label{eq:kernel_mr_coefficients} 
    \mathbf K \boldsymbol{\gamma}_{k,c}^{\mathrm{MR}} = \mathbf r_{k,c}^{\mathrm{MR}}. 
\end{align} 
When numerical regularization is required, the coefficients may instead be obtained from 
\begin{align}\label{eq:kernel_regularized_projection}
    \left( \mathbf K+\lambda_K\mathbf I \right) \boldsymbol{\gamma}_{k,c}^{\mathrm{MR}} = \mathbf r_{k,c}^{\mathrm{MR}}, \quad \lambda_K>0. 
\end{align}

\paragraph{Kernel Score-Matching Representative.}

The kernel SMR is constructed by matching the class-wise lifted gradient. From \eqref{eq:kernel_lifted_hinge}, the derivatives of a single lifted observation are 
\begin{align}
    \frac{\partial\ell_i}{\partial b} &= -c\Phi(a_i), \label{eq:kernel_gradient_b}\\ 
    \nabla_{\boldsymbol{\beta}}\ell_i &= -c\Phi(a_i)\mathbf{k}(\mathbf{x}_i) + \phi(a_i) \frac{ \mathbf C_{k,c}\boldsymbol{\beta} }{ s_{k,c} }. \label{eq:kernel_gradient_beta} 
\end{align}
Matching the intercept derivative gives 
\begin{align}\label{eq:kernel_smr_bias} 
    \Phi(\tilde a_{k,c}) = \frac{1}{n_{k,c}} \sum_{i\in\mathcal I_{k,c}} \Phi(a_i), 
\end{align}
and hence 
\begin{align}\label{eq:kernel_smr_atilde}  
    \tilde a_{k,c} = \Phi^{-1} \left\{ \frac{1}{n_{k,c}} \sum_{i\in\mathcal I_{k,c}} \Phi(a_i) \right\}. 
\end{align}

As in the linear construction, restrict the representative covariance to be proportional to the group covariance, 
\begin{align}\label{eq:kernel_smr_covariance} 
    \tilde{\mathbf C}_{k,c} = \delta_{C,k,c}\mathbf C_{k,c}, \quad \delta_{C,k,c}>0. 
\end{align}
Then 
\begin{align*} 
    \tilde s_{k,c} = \sqrt{\delta_{C,k,c}}\,s_{k,c}.
\end{align*} 
Define 
\begin{align*}
    \mathbf p_{k,c} &= \frac{1}{n_{k,c}} \sum_{i\in\mathcal I_{k,c}} \Phi(a_i)\mathbf{k}(\mathbf{x}_i),\\ \bar\phi_{k,c} &= \frac{1}{n_{k,c}} \sum_{i\in\mathcal I_{k,c}} \phi(a_i). 
\end{align*}
Matching the $\boldsymbol{\beta}$-gradient gives 
\begin{align}\label{eq:kernel_smr_gradient_matching} 
    -c\Phi(\tilde a_{k,c})\tilde{\mathbf r}_{k,c} + \phi(\tilde a_{k,c}) \sqrt{\delta_{C,k,c}} \frac{ \mathbf C_{k,c}\boldsymbol{\beta} }{ s_{k,c} } = -c\mathbf p_{k,c} + \bar\phi_{k,c} \frac{ \mathbf C_{k,c}\boldsymbol{\beta} }{ s_{k,c} }. 
\end{align}
Therefore, 
\begin{align}\label{eq:kernel_smr_interaction}
    \tilde{\mathbf r}_{k,c} = \frac{\mathbf p_{k,c}} {\Phi(\tilde a_{k,c})} + \frac{c}{\Phi(\tilde a_{k,c})} \left[ \phi(\tilde a_{k,c}) \sqrt{\delta_{C,k,c}} - \bar\phi_{k,c} \right] \frac{ \mathbf C_{k,c}\boldsymbol{\beta} }{ s_{k,c} }. 
\end{align}

The covariance scale follows from the same projected score-matching identity as in the linear case. Define $h(a)=a\Phi(a)+\phi(a)$. Then 
\begin{align}\label{eq:kernel_smr_scale}
    \sqrt{\delta_{C,k,c}} = \frac{ n_{k,c}^{-1} \sum_{i\in\mathcal I_{k,c}} h(a_i) }{ h(\tilde a_{k,c}) }, 
\end{align} 
or equivalently 
\begin{align}\label{eq:kernel_smr_delta}  
    \delta_{C,k,c} = \left[ \frac{ n_{k,c}^{-1} \sum_{i\in\mathcal I_{k,c}} h(a_i) }{ h(\tilde a_{k,c}) } \right]^2. 
\end{align}

Finally, the representative coefficients are obtained from 
\begin{align}
    \left( \mathbf K+\lambda_K\mathbf I \right) \boldsymbol{\gamma}_{k,c}^{\mathrm{SMR}} = \tilde{\mathbf r}_{k,c}. \label{eq:kernel_smr_coefficients} 
\end{align}
Thus, rather than communicating an element of the RKHS directly, node $k$ communicates the finite-dimensional tuple 
\begin{align*}
    \left( n_{k,c}, c, \boldsymbol{\gamma}_{k,c}, \tilde{\mathbf C}_{k,c} \right). 
\end{align*}
Because all nodes use the same dictionary, the received representative can be evaluated under a recipient's current parameter $\boldsymbol{\beta}$ through $\boldsymbol{\beta}^{\top}\mathbf K\boldsymbol{\gamma}_{k,c}$ and $\boldsymbol{\beta}^{\top}\tilde{\mathbf C}_{k,c}\boldsymbol{\beta}$.

\subsubsection{Distributed Representative Learning}

The distributed kernel procedure follows the same representative-exchange structure as the linear case. At communication round $t$, node $k$ constructs class-wise kernel representatives 
\begin{align*} 
\tilde{\mathcal D}_{k,c}^{(t)} = \left( n_{k,c}, c, \boldsymbol{\gamma}_{k,c}^{(t)}, \tilde{\mathbf C}_{k,c}^{(t)} \right) 
\end{align*} 
and communicates them to the other participating nodes. Kernel MR representatives are model-independent, whereas kernel SMR representatives depend on the current coefficient vector and are reconstructed as the model evolves.

After receiving the representatives, node $k$ minimizes the representative-augmented kernel objective 
\begin{align} 
    \tilde{\mathcal L}_k^{(t)}(\boldsymbol{\beta},b) &= \frac{1}{2} \boldsymbol{\beta}^{\top} \mathbf K \boldsymbol{\beta} + \frac{C}{n} \Bigg[ \sum_{c\in\{-1,1\}} \sum_{i\in\mathcal I_{k,c}} \ell_{i,k}(\boldsymbol{\beta},b) \nonumber\\
    &\hspace{2cm} + \sum_{h\neq k} \sum_{c\in\{-1,1\}} n_{h,c} \tilde\ell \left( \tilde{\mathcal D}_{h,c}^{(t)}; \boldsymbol{\beta},b \right) \Bigg], \label{eq:kernel_distributed_objective}
\end{align} 
where the lifted loss of a received representative is 
\begin{align}\label{eq:kernel_representative_loss} 
    \tilde\ell \left( \tilde{\mathcal D}_{h,c}^{(t)}; \boldsymbol{\beta},b \right) = \left( 1-\tilde\nu_{h,c}^{(t)} \right) \Phi\left( \tilde a_{h,c}^{(t)} \right) + \tilde s_{h,c}^{(t)} \phi\left( \tilde a_{h,c}^{(t)} \right), 
\end{align}
with 
\begin{align*} 
    \tilde\nu_{h,c}^{(t)} &= c\left( \boldsymbol{\beta}^{\top} \mathbf K \boldsymbol{\gamma}_{h,c}^{(t)} +b \right),\\ 
    \tilde s_{h,c}^{(t)} &= \sqrt{ \boldsymbol{\beta}^{\top} \tilde{\mathbf C}_{h,c}^{(t)} \boldsymbol{\beta} },\\ 
    \tilde a_{h,c}^{(t)} &= \frac{ 1-\tilde\nu_{h,c}^{(t)} }{ \tilde s_{h,c}^{(t)} }. 
\end{align*}

The representatives remain fixed during the local optimization phase, and node $k$ performs updates in $(\boldsymbol{\beta},b)$ analogously to the linear procedure. Consequently, the extension from the linear to the kernel setting preserves the same communication structure: nodes exchange weighted class-wise clouded representatives rather than raw observations or model parameters. The shared dictionary provides the additional common coordinate system required to express and evaluate these representatives across nodes.

\section{Numerical Experiments}

\subsection{Linear SVM Experiments}

\subsubsection{Effect of Lifting}\label{sec:effect_of_lifting}

We first isolate the effect of Gaussian lifting from the effect of representative sharing. Specifically, we compare optimization of the ordinary hinge-loss objective with optimization of the lifted hinge-loss objective on the same single-node datasets. This comparison addresses two questions: whether lifting materially changes the resulting classification performance and whether the smooth lifted objective provides more favorable behavior under comparable first-order optimization methods.

For the ordinary hinge-loss objective, we use subgradient descent with the decreasing step-size schedule 
\begin{align} 
\eta_t = \frac{\eta_0}{1+t}. \label{eq:hinge_step_schedule} 
\end{align} 
The initial step size is selected from $\eta_0 \in \{ 0.001,\, 0.003,\, 0.01,\, 0.03,\, 0.1,\, 0.3 \}$ separately for each simulation setting. The lifted objective is optimized using gradient descent with a fixed step size. These choices are intended to compare the two objectives using broadly comparable first-order procedures rather than specialized SVM solvers.

For each method, optimization performance is measured through the normalized best-so-far objective gap 
\begin{align}\label{eq:normalized_objective_gap} 
G_t = \frac{ \displaystyle \min_{0\leq s\leq t} F(\boldsymbol{\theta}^{(s)}) - F^* }{ F(\boldsymbol{\theta}^{(0)}) - F^* }, 
\end{align} 
where $F$ denotes the objective being optimized and $F^*$ is its minimum value. Because the ordinary and lifted losses define different objectives, the normalized gap is computed relative to the optimum of the corresponding objective. We record the number of iterations required to reach $G_t\leq10^{-3}$,  together with the final classification accuracy.

The experiments consider three broad simulation regimes: standard balanced samples, balanced samples with fewer observations, and samples with substantial class imbalance. Within these regimes, the data configurations vary in class separation and covariate dispersion. Table~\ref{tab:effect_of_lifting} summarizes the resulting optimization and classification performance.

\begin{table}[t] 
\centering 
\caption{ Optimization and classification performance of ordinary and lifted hinge-loss objectives. Iterations report the number of first-order updates required to reach a normalized objective gap below $10^{-3}$. } 
\label{tab:effect_of_lifting} 
\small 
\begin{tabular}{llcc} 
\toprule 
Regime & Objective & Iterations to $10^{-3}$ & Final accuracy \\ 
\midrule 
Standard & Lifted hinge & 22.5 & 0.9353 \\
& Ordinary hinge & 81.4 & 0.9356 \\ 
\midrule 
Few balanced & Lifted hinge & 24.9 & 0.9648 \\ 
& Ordinary hinge & 130.1 & 0.9648 \\ 
\midrule
Heavy imbalance & Lifted hinge & 46.2 & 0.9765 \\ 
& Ordinary hinge & 81.4 & 0.9771 \\ 
\bottomrule 
\end{tabular} 
\end{table}

The ordinary and lifted hinge-loss objectives yield nearly identical final classification accuracies in all three reported regimes. Thus, within these simulations, replacing the pointwise hinge loss by its Gaussian-lifted counterpart does not materially alter predictive performance. This observation is consistent with the limiting result in Proposition~\ref{prop:degen_liftedloss}, although the simulations use nondegenerate lifting covariances.

The principal difference appears in optimization. Under the first-order methods considered here, gradient descent on the smooth lifted objective reaches the specified objective-gap tolerance in fewer iterations than subgradient descent on the ordinary hinge-loss objective. The difference is largest in the few-observation balanced regime, where the lifted objective requires 24.9 iterations on average, compared with 130.1 for the ordinary hinge objective.

This comparison should not be interpreted as establishing that gradient descent on the lifted objective is uniformly faster than specialized SVM optimization methods. Rather, it illustrates the computational effect of replacing a nonsmooth objective by a differentiable expected loss when comparable first-order procedures are used. The smooth gradient structure is particularly relevant to the clouded SMR construction, which uses local gradient information to construct and update representatives.

\subsubsection{Distributed Learning under Heterogeneity} 

For the distributed experiments that follow, we use $100$ independent replications for each configuration. Each dataset contains $600$ observations in dimensions $d\in\{2,3,5,7\}$,  with equal global proportions of the two classes before distributed partitioning. The class-conditional distributions are three-component Gaussian mixtures. For each class $c\in\{-1,1\}$, the mixture weights are sampled according to $\left( \pi_{c,1}, \pi_{c,2}, \pi_{c,3} \right) \sim \operatorname{Dirichlet}(1,1,1)$,  and the component means are generated as  $\boldsymbol{\mu}_{c,j} = \boldsymbol{\mu}_c + r_{c,j}\mathbf z_{c,j}$, $r_{c,j} \sim \operatorname{Unif}(0,0.75)$,  where $\mathbf z_{c,j}$ is a randomly generated unit vector. Let $m$ denote one-half of the distance between the two base class centers. The high-overlap configurations use $m\sim\operatorname{Unif}(0.25,1.00)$,  whereas the low-overlap configurations use $m\sim\operatorname{Unif}(1.00,2.50)$. 

The Gaussian-component covariance matrices are selected independently for the two classes from a covariance bank containing dispersion levels $0.25$, $0.50$, $1.00$, $1.50$, $2.00$.  At each dispersion level, the bank contains isotropic, mildly anisotropic, and strongly anisotropic covariance structures. We write the generating covariance as 
\begin{align*}
    \boldsymbol{\Omega}_{c} = s_c^2\mathbf A_c, \quad \frac{\operatorname{tr}(\mathbf A_c)}{d}=1, 
\end{align*}
so that 
\begin{align*}
    \sqrt{ \frac{ \operatorname{tr}(\boldsymbol{\Omega}_{c}) }{d} } = s_c.
\end{align*}
For the isotropic structure, $\mathbf A_c=\mathbf I_d$. Before trace normalization, the mildly and strongly anisotropic structures assign, respectively, two and twenty-five times as much variance to the first coordinate. The generating covariances for the two classes are selected independently. Independent test observations are generated from the same class-conditional Gaussian-mixture distributions as the corresponding training observations.

The observations are distributed across three nodes using independent class-wise Dirichlet-multinomial allocations with concentration parameter $\alpha=0.1$. The allocation is redrawn until every node contains at least two observations from each class. This procedure preserves global class balance while producing strongly heterogeneous local class proportions.

For each node-class group $(k,c)$, the lifting covariance is estimated from the corresponding local observations. Let $\widehat{\sigma}_{k,c,j}$ denote the empirical standard deviation of the $j$th covariate among the $n_{k,c}$ observations in that group. We use the bandwidth factor \[ h_{k,c} = \left( \frac{3n_{k,c}}{4} \right)^{-1/5} \] and define 
\begin{align} 
    \boldsymbol{\Sigma}_{k,c} = \operatorname{diag} \left[ \left( h_{k,c}\widehat{\sigma}_{k,c,1} \right)^2, \ldots, \left( h_{k,c}\widehat{\sigma}_{k,c,d} \right)^2 \right]. \label{eq:linear_experiment_covariance} 
\end{align} 
When this covariance cannot be estimated reliably, it is replaced by a small positive-definite diagonal matrix to maintain numerical stability.

All distributed methods use the same initialization, $\mathbf w^{(0)}=\mathbf 0$,  $b^{(0)}=0$,  followed by $20$ local warm-start iterations before the first communication round. Unless otherwise stated, the distributed methods use learning rate $0.005$, regularization parameter $C=75$ with the loss normalized by the total sample size, $40$ local optimization steps per communication round, and $25$ communication rounds.

We consider both low- and high-overlap regimes. Since the high-overlap setting provides the more difficult classification problem and the stronger test of optimization under heterogeneous local class proportions, we report its predictive-accuracy results below. Both regimes are retained in the subsequent comparison of recovery of the centralized solution.

We compare the proposed clouded representatives with parameter-based and centralized alternatives. MR (Lifted) and SMR (Lifted) use the lifted hinge loss developed in Section~\ref{sec:linear_rep}. Their nonsmooth counterparts, MR (Hinge) and SMR (Hinge), use the ordinary hinge objective. We additionally include FedAvg and FedProx using the lifted loss, together with a distributed ADMM solution for the ordinary hinge objective. Two centralized benchmarks are included: Classical Linear SVM, which fits the ordinary hinge-loss objective to the pooled data, and Global Pooled (Lifted), which minimizes the lifted objective using all observations.

\begin{table}[t]
\centering 
\caption{Test accuracy over $100$ simulations for high-overlap classes under the default communication setting. Values are mean $\pm$ standard deviation.}
\label{tab:linear_test_accuracy_overlap}
{\footnotesize
\begin{tabular}{lcccc} 
\toprule 
Method & $d=2$ & $d=3$ & $d=5$ & $d=7$ \\ 
\midrule
MR (Hinge) & $0.7599\pm0.0982$ & $0.7811\pm0.0920$ & $0.7866\pm0.0968$ & $0.7834\pm0.0913$ \\ 
SMR (Hinge) & $0.6653\pm0.1474$ & $0.6838\pm0.1497$ & $0.6872\pm0.1544$ & $0.6746\pm0.1391$ \\ 
SMR (Lifted) & $0.7545\pm0.1049$ & $0.7860\pm0.0910$ & $\mathbf{0.7995\pm0.1052}$ & $0.7879\pm0.0965$ \\
MR (Lifted) & $\mathbf{0.7601\pm0.1003}$ & $\mathbf{0.7873\pm0.0913}$ & $0.7990\pm0.1039$ & $\mathbf{0.7938\pm0.0942}$ \\
FedAvg (Lifted) & $0.7345\pm0.1258$ & $0.7478\pm0.1221$ & $0.7650\pm0.1241$ & $0.7417\pm0.1289$ \\ 
FedProx (Lifted) & $0.7356\pm0.1191$ & $0.7469\pm0.1251$ & $0.7653\pm0.1252$ & $0.7427\pm0.1266$ \\ 
\midrule ADMM (Hinge) & $0.7437\pm0.1223$ & $0.7675\pm0.1124$ & $0.7833\pm0.1308$ & $0.7819\pm0.1017$ \\
Classical SVM (Hinge) & $0.7482\pm0.1126$ & $0.7861\pm0.0933$ & $0.7981\pm0.1044$ & $0.7911\pm0.0952$ \\ 
Global Pooled (Lifted) & $0.7539\pm0.1044$ & $0.7868\pm0.0909$ & $\mathbf{0.8005\pm0.1057}$ & $0.7895\pm0.0950$ \\ 
\bottomrule
\end{tabular} 
}
\end{table}

Under the high-overlap setting, the effect of heterogeneous local class proportions is visible in the parameter-aggregation methods. With $40$ local updates between communications, FedAvg and FedProx have lower test accuracy than the centralized lifted solution across all four dimensions. In contrast, the two lifted representative methods remain close to the centralized benchmarks. In particular, MR (Lifted) and SMR (Lifted) produce test accuracies comparable to Global Pooled (Lifted) throughout the reported dimensions.

The comparison between SMR (Hinge) and SMR (Lifted) is also notable. Score matching based directly on the nonsmooth hinge objective performs substantially worse in these experiments, whereas clouded SMR constructed from the smooth lifted objective remains competitive with the centralized solutions. The corresponding difference between MR (Hinge) and MR (Lifted) is considerably smaller. This distinction is consistent with the role of lifting in the two constructions: MR is primarily a model-independent summary of class-wise location, while SMR directly uses objective-gradient information when constructing its representatives.

\subsubsection{Communication Frequency} 

The preceding experiment uses $40$ local gradient steps between representative exchanges. We next vary the communication frequency to examine how long communicated information remains useful as the local models move away from the parameter values at which it was constructed. We focus on the high-overlap setting, where local class imbalance and overlap provide the stronger test of distributed optimization.

We consider three communication regimes. The \emph{moderate} regime uses $25$ communication rounds with $40$ local gradient steps per round. The \emph{single-communication} regime performs only one representative exchange before extensive subsequent local optimization. Finally, the \emph{frequent-communication} regime takes one local gradient step between successive communication rounds. These settings distinguish methods that depend strongly on the current model state from summaries that remain fixed throughout optimization.

In addition to predictive accuracy, we examine whether the local models agree after distributed training. For a given simulation, model agreement is declared when the maximum Euclidean parameter distance between any two nodes is less than $10^{-6}$. Table~\ref{tab:linear_smr_communication_agreement} reports the percentage of simulations satisfying this criterion for lifted SMR.

\begin{table}[t] 
\centering 
\caption{Percentage of high-overlap simulations in which the local lifted-SMR models agree under different communication regimes.}
\label{tab:linear_smr_communication_agreement}
\small 
\begin{tabular}{lcccc} 
\toprule 
Communication regime & $d=2$ & $d=3$ & $d=5$ & $d=7$ \\ 
\midrule 
Moderate & $99\%$ & $100\%$ & $100\%$ & $95\%$ \\
Single communication & $0\%$ & $0\%$ & $0\%$ & $0\%$ \\
Frequent & $100\%$ & $100\%$ & $100\%$ & $96\%$ \\
\bottomrule 
\end{tabular} 
\end{table}

The communication regimes reveal a clear distinction between MR and SMR. MR is determined by the local data and is independent of the current model parameter. Its representative therefore remains unchanged as local optimization proceeds. SMR instead matches first-order information at the parameter value where the representative is constructed. When many local updates occur without reconstruction, the local parameters can move away from this construction point and the frozen representative becomes a less accurate description of the current local gradient.

This behavior is most apparent in the single-communication regime. Under moderate communication, lifted SMR achieves model agreement in $95\%$-$100\%$ of the simulations across the four dimensions, and under frequent communication the corresponding values are $96\%$-$100\%$. When the representatives are communicated only once, however, agreement falls to zero in all four dimensions. Thus, the SMR should be interpreted as a locally accurate, periodically refreshed representation of the optimization objective rather than as a static summary of the local data.

When communication is frequent and only one local update is performed between representative reconstructions, the distinction between distributed methods is correspondingly reduced. This behavior is also consistent with the exact score-matching property discussed in Remark~\ref{rem:exact_first_step}: immediately after an SMR reconstruction, the first representative-gradient update agrees with the corresponding full-data gradient update at the construction parameter. The empirical results therefore illustrate the communication-local-optimization trade-off predicted by the frozen-representative analysis.

\subsubsection{Recovery of the Centralized Solution}

Predictive accuracy and agreement among nodes do not, by themselves, establish that a distributed method has recovered the centralized optimizer. Different linear classifiers can produce similar predictions while having substantially different parameter values. We therefore compare the representative-based solutions directly with the Global Pooled (Lifted) solution.

This comparison uses the default moderate-communication setting. For each method, we measure the Euclidean discrepancy between the learned weight vector and the centralized lifted-SVM weight vector. Table~\ref{tab:linear_parameter_distance} reports the resulting distances for both the low- and high-overlap regimes. The corresponding comparisons of intercept discrepancy and cosine similarity lead to the same qualitative conclusion and are therefore omitted for brevity.

\begin{table}[t]
\centering 
\caption{Euclidean distance $\|\mathbf w-\mathbf w_{\mathrm{global}}\|_2$ from the Global Pooled (Lifted) solution under the default communication setting.} 
\label{tab:linear_parameter_distance} 
\small 
\begin{tabular}{llcccc}
\toprule 
Setting & Method & $d=2$ & $d=3$ & $d=5$ & $d=7$ \\ 
\midrule
Low overlap & SMR (Lifted) & $\mathbf{0.0444}$ & $\mathbf{0.0457}$ & $\mathbf{0.0618}$ & $\mathbf{0.0651}$ \\
& SMR (Hinge) & $0.4497$ & $0.5059$ & $0.5156$ & $0.5498$ \\
& MR (Lifted) & $0.6032$ & $0.5946$ & $0.5951$ & $0.5983$ \\ 
& MR (Hinge) & $0.7943$ & $0.7335$ & $0.7337$ & $0.7578$ \\ 
\midrule 
High overlap & SMR (Lifted) & $\mathbf{0.0475}$ & $\mathbf{0.0485}$ & $\mathbf{0.0699}$ & $\mathbf{0.0714}$ \\ 
& SMR (Hinge) & $0.5200$ & $0.5906$ & $0.5872$ & $0.6229$ \\ & MR (Lifted) & $0.6860$ & $0.6658$ & $0.6641$ & $0.6632$ \\
& MR (Hinge) & $0.8847$ & $0.8094$ & $0.8059$ & $0.8312$ \\ 
\bottomrule 
\end{tabular} 
\end{table}

The parameter-distance comparison separates the roles of MR and SMR more clearly than predictive accuracy alone. Across every reported dimension and in both overlap regimes, lifted SMR produces the smallest discrepancy from the centralized lifted-SVM weight vector. Its distances range from $0.0444$ to $0.0651$ in the low-overlap setting and from $0.0475$ to $0.0714$ in the high-overlap setting. By comparison, the lifted-MR distances are approximately $0.59$-$0.60$ under low overlap and $0.66$-$0.69$ under high overlap.

This distinction helps explain why MR and SMR can have similar classification accuracy while behaving differently as optimization procedures. MR preserves class-wise data location and can therefore provide a strong predictive summary in these approximately linear settings, but it is not constructed to reproduce the local objective gradient. SMR is explicitly constructed to match that first-order information. The substantially smaller parameter discrepancies for lifted SMR therefore provide direct empirical evidence that score matching more faithfully recovers the optimizer of the centralized lifted objective.

Taken together, the three distributed experiments distinguish predictive performance, communication sensitivity, and optimization fidelity. Clouded MR provides a stable model-independent summary that can remain useful under infrequent communication, while clouded SMR is model-dependent and benefits from periodic reconstruction. When refreshed during optimization, lifted SMR not only maintains competitive predictive accuracy under heterogeneous local class distributions but also most closely reproduces the centralized lifted-SVM solution.

\subsection{Kernel SVM Experiments}

We next evaluate the kernel extension on two binary-classification datasets using the Gaussian RBF kernel. In contrast to the synthetic experiments used for the linear SVM, these experiments use the Ionosphere and Australian Credit Approval datasets and evaluate the representative methods under heterogeneous class allocation across nodes.

\subsubsection{Experimental Setup}

We evaluate the kernel extension on the Ionosphere and Australian Credit Approval binary-classification datasets using the Gaussian RBF kernel. All reported results are averaged over $100$ independent repetitions.

For each repetition, the observations are randomly divided into $85\%$ training data and $15\%$ test data. Covariate standardization is fitted using only the training data, and the resulting transformation is then applied to the private training observations, the shared dictionary, and the held-out test observations. Thus, the test data are not used to estimate the preprocessing transformation.

The kernel representative construction requires a common dictionary through which all nodes express their models and representatives. After standardization, a stratified random sample containing $10\%$ of the training observations is selected as the shared dictionary. These observations are removed from the private training pool before the remaining observations are distributed across the nodes.

The Ionosphere dataset contains $351$ observations, yielding $298$ training observations and $53$ test observations. Its shared dictionary contains $30$ observations. The Australian Credit Approval dataset contains $690$ observations, yielding $586$ training observations and $104$ test observations. Its shared dictionary contains $59$ observations.

After removing the dictionary observations, the remaining private training data are distributed across three nodes using independent class-wise Dirichlet allocations with concentration parameter $\alpha=0.1$. The allocation is redrawn until each node contains at least four observations in total and at least two observations from each class.

All kernel representative methods use the Gaussian RBF kernel 
\begin{align} 
K(\mathbf x,\mathbf x') = \exp \left( -\gamma \| \mathbf x-\mathbf x' \|^2 \right), \quad \gamma=\frac{1}{d}, \label{eq:kernel_experiment_rbf} 
\end{align} 
where $d$ is the covariate dimension of the corresponding dataset. The distributed lifted methods use regularization parameter $C=100$, with the empirical-loss term normalized by the total sample size. The centralized classical kernel-SVM baseline uses $C=1$.

Before the first representative exchange, each node performs $50$ local warm-start optimization steps with learning rate $0.001$. For the hinge-loss first-order methods, we use initial step size $0.1$ together with an inverse-square-root step-size schedule.

As in the linear experiments, we compare three communication regimes. We denote by $J/T$ a regime with $J$ local optimization steps per communication round and $T$ communication rounds, and consider $1/1000$,  $40/25$, $1000/1$.  These correspond, respectively, to frequent communication, moderate communication, and a single communication round followed by extensive local optimization.

We compare lifted MR and SMR with their hinge-loss counterparts, FedAvg and FedProx using the lifted objective, and distributed ADMM using the ordinary hinge loss. We additionally report three centralized reference solutions. Kernel SVM (Hinge) fits the ordinary RBF-kernel SVM. Central Full (Lifted) minimizes the lifted kernel objective using the full training-set dictionary, whereas Central Shared (Lifted) minimizes the lifted objective in the smaller dictionary span available to the distributed methods. Central Shared (Lifted) therefore provides the most direct centralized target for evaluating the distributed representative methods.

\subsubsection{Distributed Learning and Communication Frequency}

Table~\ref{tab:kernel_test_accuracy} reports test accuracy under the three communication regimes. Because the distributed kernel methods are restricted to the common dictionary span, both centralized full-dictionary and shared-dictionary results are included. Their difference separates the approximation introduced by restricting the kernel model to the shared dictionary from the additional approximation introduced by distributed learning.

\begin{table*}[t] 
\centering 
\caption{ Test accuracy over $100$ repetitions for the kernel SVM experiments. Columns denote local steps per communication round / number of communication rounds. Values are mean $\pm$ standard deviation. } 
\label{tab:kernel_test_accuracy}
\footnotesize 
\begin{tabular}{llccc} 
\toprule 
Dataset & Method & $1/1000$ & $40/25$ & $1000/1$ \\ 
\midrule
Ionosphere & MR (Hinge) & $0.8550\pm0.0541$ & $0.8550\pm0.0541$ & $0.8550\pm0.0541$ \\ 
& SMR (Hinge) & $0.9189\pm0.0398$ & $0.9032\pm0.0499$ & $0.7473\pm0.1640$ \\ 
& SMR (Lifted) & $\mathbf{0.9276\pm0.0373}$ & $\mathbf{0.9277\pm0.0376}$ & $0.9101\pm0.0486$ \\
& MR (Lifted) & $0.9233\pm0.0391$ & $0.9233\pm0.0391$ & $\mathbf{0.9233\pm0.0391}$ \\ 
& FedAvg (Lifted) & $0.9275\pm0.0372$ & $0.8855\pm0.0810$ & $0.8408\pm0.1180$ \\ 
& FedProx (Lifted) & $0.9275\pm0.0372$ & $0.8847\pm0.0819$ & $0.8387\pm0.1168$ \\
& ADMM (Hinge) & $0.9200\pm0.0392$ & $0.9204\pm0.0401$ & $0.7843\pm0.1454$ \\ 
& Kernel SVM (Hinge) & \multicolumn{3}{c}{$0.9402\pm0.0325$} \\ 
& Central Full (Lifted) & \multicolumn{3}{c}{$0.9383\pm0.0324$} \\
& Central Shared (Lifted) & \multicolumn{3}{c}{$0.9309\pm0.0342$} \\ 
\midrule 
Australian & MR (Hinge) & $0.8547\pm0.0300$ & $0.8547\pm0.0300$ & $0.8547\pm0.0300$ \\
& SMR (Hinge) & $0.8572\pm0.0287$ & $0.8424\pm0.0442$ & $0.7178\pm0.1686$ \\ 
& SMR (Lifted) & $\mathbf{0.8658\pm0.0313}$ & $\mathbf{0.8657\pm0.0312}$ & $0.8605\pm0.0319$ \\ 
& MR (Lifted) & $0.8657\pm0.0322$ & $\mathbf{0.8657\pm0.0322}$ & $\mathbf{0.8657\pm0.0322}$ \\
& FedAvg (Lifted) & $\mathbf{0.8658\pm0.0314}$ & $0.8158\pm0.0876$ & $0.7917\pm0.1137$ \\
& FedProx (Lifted) & $\mathbf{0.8658\pm0.0314}$ & $0.8151\pm0.0893$ & $0.7964\pm0.1066$ \\ 
& ADMM (Hinge) & $0.8618\pm0.0330$ & $0.8617\pm0.0323$ & $0.7264\pm0.1532$ \\
&   Kernel SVM (Hinge) & \multicolumn{3}{c}{$0.8586\pm0.0310$} \\ 
& Central Full (Lifted) & \multicolumn{3}{c}{$0.8668\pm0.0308$} \\
& Central Shared (Lifted) & \multicolumn{3}{c}{$0.8671\pm0.0308$} \\
\bottomrule 
\end{tabular}
\end{table*}

The centralized comparisons show that the effect of restricting the model to the shared dictionary depends on the dataset. On Ionosphere, the Central Shared Dictionary solution has test accuracy $0.9309$, compared with $0.9383$ for the full-dictionary lifted solution. On the Australian dataset, the corresponding accuracies are $0.8671$ and $0.8668$, respectively. Thus, the shared-dictionary restriction produces a visible accuracy difference for Ionosphere but essentially no difference for Australian in these experiments.

Under frequent communication, the primary distributed lifted methods are close to the Central Shared Dictionary solution on both datasets. As the number of local updates between communications increases, however, the parameter-aggregation methods become substantially more sensitive to local drift. For example, on Ionosphere, FedAvg decreases from $0.9275$ under the $1/1000$ regime to $0.8855$ under $40/25$ and $0.8408$ under $1000/1$. The corresponding Australian accuracies are $0.8658$, $0.8158$, and $0.7917$.

Lifted MR and lifted SMR are considerably less sensitive to the moderate-communication regime. On Ionosphere, their test accuracies under $40/25$ are $0.9233$ and $0.9277$, respectively, compared with $0.9309$ for the centralized shared-dictionary objective. On Australian, both representative methods obtain $0.8657$, compared with $0.8671$ for the centralized shared-dictionary solution.

The single-communication regime also illustrates the different roles of MR and SMR. MR is independent of the current model and therefore remains unchanged as local optimization proceeds. SMR instead matches the objective gradient at its construction point and becomes less representative as the model moves away from that point without reconstruction. Accordingly, lifted MR is unchanged across the three communication regimes, while lifted SMR exhibits some degradation under $1000/1$. Nevertheless, lifted SMR remains substantially more stable in this regime than both FedAvg and the hinge-based SMR construction.

To further examine the effect of representative refresh frequency, we measure agreement among the local lifted-SMR models. Agreement is defined using an RKHS-distance tolerance of $10^{-2}$.

\begin{table}[t] 
\centering
\caption{ Percentage of repetitions in which the local lifted-SMR kernel models agree within RKHS-distance tolerance $10^{-2}$. } \label{tab:kernel_model_agreement} 
\small 
\begin{tabular}{lccc} \toprule Dataset & $1/1000$ & $40/25$ & $1000/1$ \\ 
\midrule
Ionosphere & $90\%$ & $84\%$ & $0\%$ \\ Australian & $93\%$ & $93\%$ & $0\%$ \\
\bottomrule 
\end{tabular}
\end{table}

Lifted SMR achieves agreement in $84\%$--$93\%$ of repetitions when representatives are refreshed during training, but no repetitions satisfy the agreement criterion when only one representative exchange is performed. This behavior parallels the linear experiments and reinforces the interpretation of SMR as a local, model-dependent approximation that benefits from periodic reconstruction.

\subsubsection{Recovery of the Centralized Solution}

Classification accuracy does not determine whether a distributed classifier has recovered the centralized optimization solution. We therefore compare each representative-based model directly with the Central Shared Dictionary (Lifted) solution. Because all models are expressed in the same dictionary span, the discrepancy can be measured using the RKHS norm.

Let $w=\Psi_{\mathcal B}\boldsymbol{\beta}$, $w_g=\Psi_{\mathcal B}\boldsymbol{\beta}_g$  denote a distributed model and the corresponding centralized shared-dictionary model. Their RKHS distance is 
\begin{align}\label{eq:kernel_experiment_rkhs_distance} 
\|w-w_g\|_{\mathcal H} = \sqrt{ (\boldsymbol{\beta}-\boldsymbol{\beta}_g)^\top \mathbf K (\boldsymbol{\beta}-\boldsymbol{\beta}_g) }. 
\end{align} 
For each repetition, the distance is first averaged across the three local node models and then averaged across repetitions.

\begin{table*}[t]
\centering
\caption{ RKHS distance from the centralized shared-dictionary lifted solution. Values are mean $\pm$ standard deviation. }
\label{tab:kernel_rkhs_distance} 
\small 
\begin{tabular}{llccc} 
\toprule 
Dataset & Method & $1/1000$ & $40/25$ & $1000/1$ \\ 
\midrule 
Ionosphere & MR (Hinge) & $2.0588\pm0.2334$ & $2.0588\pm0.2334$ & $2.0588\pm0.2334$ \\
& MR (Lifted) & $0.7246\pm0.1988$ & $0.7246\pm0.1988$ & $0.7246\pm0.1988$ \\ 
& SMR (Hinge) & $0.7684\pm0.2062$ & $0.9086\pm0.2593$ & $2.0234\pm0.7213$ \\ 
& SMR (Lifted) & $\mathbf{0.5961\pm0.2039}$ & $\mathbf{0.5960\pm0.2038}$ & $\mathbf{0.6972\pm0.2386}$ \\
\midrule 
Australian & MR (Hinge) & $1.9901\pm0.1043$ & $1.9901\pm0.1043$ & $1.9901\pm0.1043$ \\
& MR (Lifted) & $0.5755\pm0.0422$ & $0.5755\pm0.0422$ & $0.5755\pm0.0422$ \\ 
& SMR (Hinge) & $0.6300\pm0.1177$ & $0.8669\pm0.1938$ & $1.8893\pm0.6440$ \\ 
& SMR (Lifted) & $\mathbf{0.2967\pm0.0667}$ & $\mathbf{0.2970\pm0.0666}$ & $\mathbf{0.4642\pm0.1331}$ \\ 
\bottomrule 
\end{tabular} 
\end{table*}

The RKHS-distance results provide a sharper distinction between MR and SMR than predictive accuracy alone. Lifted MR and lifted SMR have very similar test accuracy in several settings, but lifted SMR is consistently closer to the centralized shared-dictionary solution in RKHS distance. For example, under the moderate $40/25$ regime, the mean RKHS distance on Ionosphere is $0.5960$ for lifted SMR and $0.7246$ for lifted MR. On Australian, the corresponding distances are $0.2970$ and $0.5755$.

This distinction is consistent with the representative constructions. MR summarizes class-wise location without depending on the current model, whereas SMR is explicitly constructed to reproduce first-order information from the lifted objective. Strong predictive performance from MR therefore does not necessarily imply recovery of the centralized optimizer. The RKHS-distance results show that lifted SMR more faithfully reproduces the centralized shared-dictionary kernel solution across all three communication regimes considered here.

\subsubsection{Random Fourier Feature Approximation}

As an alternative to the shared-dictionary kernel construction, we also consider a Random Fourier Feature (RFF) approximation to the RBF kernel. Whereas the shared-dictionary approach restricts the classifier to a finite-dimensional subspace generated by selected dictionary observations, RFF provides all nodes with the same finite-dimensional Euclidean feature map. The linear lifted-SVM representative constructions can then be applied directly in this common feature space.

For the RBF kernel 
\begin{align*} 
    K(\mathbf x,\mathbf x') = \exp \left( -\gamma \|\mathbf x-\mathbf x'\|^2 \right), 
\end{align*}
we sample 
\begin{align*}
\boldsymbol{\omega}_1,\ldots,\boldsymbol{\omega}_D \sim \mathcal N(\mathbf 0,2\gamma\mathbf I) 
\end{align*}
and use the feature map
\begin{align*}
\mathbf z(\mathbf x) = \frac{1}{\sqrt D} 
\begin{pmatrix}
\cos(\boldsymbol{\omega}_1^\top\mathbf x)\\ 
\sin(\boldsymbol{\omega}_1^\top\mathbf x)\\
\vdots\\ \cos(\boldsymbol{\omega}_D^\top\mathbf x)\\ 
\sin(\boldsymbol{\omega}_D^\top\mathbf x) 
\end{pmatrix}.
\end{align*} 
Then 
\begin{align*} 
    \mathbf z(\mathbf x)^\top\mathbf z(\mathbf x') \approx K(\mathbf x,\mathbf x'), 
\end{align*}
and the resulting classifier has the linear form 
\begin{align*}
    f(\mathbf x) = \mathbf w^\top\mathbf z(\mathbf x)+b. 
\end{align*}

We use $D=50$ sampled frequencies, producing a $100$-dimensional feature representation, and otherwise follow the same three-node heterogeneous partitioning and communication regimes as in the shared-dictionary kernel experiments.

\section{Conclusion and Future Work}

\subsection{Conclusion} 

We developed a Representative Learning framework for distributed support vector machines based on Lifted Empirical Risk Minimization. Rather than optimizing the nonsmooth hinge loss directly on point-valued observations, the lifted formulation replaces each covariate by a local Gaussian cloud and evaluates the expected hinge loss under this distribution. The resulting objective has an exact closed form, is differentiable for nondegenerate lifting covariances, and approaches the ordinary hinge-loss objective as the covariance scale vanishes.

The lifted formulation leads naturally to clouded representatives that communicate both location and covariance information. We considered two class-wise constructions. Clouded MR provides a model-independent summary of local class location and variability, whereas clouded SMR additionally matches the local gradient of the lifted objective at its construction parameter. These representatives are exchanged in place of raw observations, gradients, or model parameters and are incorporated directly into the local empirical-risk objectives.

Our theoretical analysis characterizes the approximation induced by frozen clouded SMR representatives. Under boundedness and nondegeneracy conditions, the representative covariance scale remains within first order of the original covariance scale as the within-group covariate diameter decreases. This structural control yields a corresponding first-order bound on the difference between the representative and full-data gradients. Under strong convexity and smoothness, the resulting iterative procedure can be viewed as inexact gradient descent, with an optimization error floor controlled by the representative approximation error. Moreover, score matching makes the first update following a synchronized representative reconstruction exact with respect to the corresponding full-data gradient.

The numerical experiments support these theoretical distinctions. For linear SVMs, the lifted objective preserves classification performance while providing a smooth objective for first-order optimization. Under heterogeneous class allocations, lifted MR and lifted SMR remain competitive with centralized solutions when parameter-aggregation methods are affected by repeated local optimization. Their different constructions become most apparent when communication is infrequent: MR remains stable as a model-independent summary, whereas SMR benefits from periodic reconstruction as the model moves away from its score-matching point. Although MR can achieve predictive performance comparable to SMR, the parameter-recovery experiments show that lifted SMR more closely reproduces the centralized lifted-SVM optimizer.

The same qualitative behavior extends to nonlinear SVMs. A shared kernel dictionary provides a common finite-dimensional coordinate system in which nodes can construct, communicate, and evaluate kernel representatives. Across the Ionosphere and Australian Credit Approval experiments, the representative methods remain competitive with the corresponding centralized shared-dictionary objectives, while RKHS model-distance comparisons again distinguish predictive performance from recovery of the centralized optimizer. Experiments using Random Fourier Features further show that representative learning can also be combined with an alternative finite-dimensional approximation of the RBF kernel.

Taken together, these results demonstrate that lifting and representative construction address complementary aspects of distributed SVM learning. Lifting supplies a differentiable objective that retains the large-margin classification structure, while clouded representatives provide compact distribution-valued summaries through which heterogeneous nodes can exchange local data and objective information. This combination offers a data-centric alternative to repeated parameter aggregation for distributed large-margin learning.

\subsection{Future Work}

Several directions follow naturally from the present framework. First, the kernel construction currently assumes a common dictionary available to all nodes. In the experiments, this dictionary is formed from a shared subset of the training data. A more fully representative-based construction would build the dictionary itself from communicated summaries rather than from shared observations. For example, class-wise MR or SMR locations could provide initial dictionary elements, while multiple representatives per class could be used when a single summary is insufficient to capture local nonlinear structure. Such a construction would remove the present reliance on shared training observations and would make both the model representation and the distributed optimization depend entirely on representative communication.

Second, the theoretical analysis can be extended beyond the synchronized, locally strongly convex setting considered here. The present gradient discrepancy result treats representatives as frozen within each local optimization phase, and the iterative convergence result assumes a common construction parameter and uniform control of the representative error. An important next step is to analyze node-specific construction parameters, where representatives may be generated from different local iterates, and to quantify explicitly how model disagreement and representative staleness enter the convergence bound. Related questions arise under asynchronous communication, partial participation, and decentralized network topologies.

Third, clouded representatives suggest a broader extension of Representative Learning from point summaries to structured probability-valued summaries. The SVM construction studied here retains local location and covariance information because both quantities enter naturally into the Gaussian-lifted hinge loss. More generally, distributional representatives could encode within-node uncertainty, multimodality, or local geometry that cannot be captured by a single pseudo-observation. Developing principled constructions for richer representative families, together with approximation guarantees that relate their complexity to local heterogeneity, is a natural direction for extending Lifted ERM beyond the SVM setting.

Finally, the distinction between model-independent and gradient-matching representatives is relevant to more expressive learning systems. A representative need not reproduce the gradient of an entire model; it may instead be constructed to preserve task-relevant information at a selected feature representation or layer. This suggests studying representative construction in learned embedding spaces and other structured feature spaces, where compact synthetic summaries could support distributed optimization without requiring direct exchange of the full model state. The present SVM results provide a tractable setting in which this principle can be studied analytically, while extensions to richer model classes remain an important direction for future work.

\bibliographystyle{chicago}
\bibliography{refer}

\end{document}